\documentclass[reqno,10pt]{amsart}
\usepackage{graphics}
\usepackage{amsmath,amssymb,amsfonts}
\usepackage{mathrsfs}
\usepackage{wasysym}
\usepackage{algorithm}
\usepackage{algorithmic}

\usepackage{pstricks}
\usepackage{smartdiagram}
\usetikzlibrary{calc,trees,positioning,arrows,chains,shapes.geometric,%
	decorations.pathreplacing,decorations.pathmorphing,shapes,%
	matrix,shapes.symbols}

\tikzset{
	>=stealth',
	punktchain/.style={
		rectangle, 
		rounded corners, 
		draw=black, very thick,
		text width=10em, 
		minimum height=2em, 
		text centered, 
		on chain},
	line/.style={draw, thick, <-},
	element/.style={
		tape,
		top color=white,
		bottom color=blue!50!black!60!,
		minimum width=6em,
		draw=blue!40!black!90, very thick,
		text width=10em, 
		minimum height=3.5em, 
		text centered, 
		on chain},
	every join/.style={->, thick,shorten >=1pt},
	decoration={brace},
	tuborg/.style={decorate},
	tubnode/.style={midway, right=2pt},
}

\tikzstyle{arrow} = [thick,->,>=stealth]

\usepackage{multirow,booktabs,comment}
\usepackage{epstopdf}

\usepackage{todonotes}
\usepackage[normalem]{ulem}

\usepackage{xcolor}
\usepackage{soul}

\usepackage[colorlinks=true,urlcolor=blue,citecolor=red,
anchorcolor=black,linkcolor=blue]{hyperref}

\usepackage{subcaption}

\usepackage{lineno}

\newtheorem{theorem}{Theorem}
\theoremstyle{plain}

\newtheorem{lemma}[theorem]{Lemma}
\newtheorem{remark}[theorem]{Remark}

\numberwithin{equation}{section}
\numberwithin{equation}{section}
\numberwithin{theorem}{section}

\newcommand{\bx}{\boldsymbol{x}}
\newcommand{\by}{\boldsymbol{y}}
\newcommand{\bX}{\boldsymbol{X}}
\newcommand{\bY}{\boldsymbol{Y}}
\newcommand{\bZ}{\boldsymbol{Z}}

\newcommand{\bs}{\boldsymbol{s}}
\newcommand{\bt}{\boldsymbol{t}}
\newcommand{\ba}{\boldsymbol{a}}
\newcommand{\bb}{\boldsymbol{b}}

\newcommand{\bz}{\boldsymbol{z}}
\newcommand{\bq}{\boldsymbol{q}}
\newcommand{\bv}{\boldsymbol{v}}
\newcommand{\bu}{\boldsymbol{u}}

\newcommand{\bxi}{\boldsymbol{\xi}}
\newcommand{\btheta}{\boldsymbol{\theta}}
\newcommand{\bphi}{\boldsymbol{\phi}}
\newcommand{\bbeta}{\boldsymbol{\beta}}
\newcommand{\balpha}{\boldsymbol{\alpha}}
\newcommand{\bmu}{\boldsymbol{\mu}}
\newcommand{\bsigma}{\boldsymbol{\sigma}}
\newcommand{\bgamma}{\boldsymbol{\gamma}}

\newcommand{\mbI}{\mathbf{I}}

\newcommand{\mbbR}{\mathbb{R}}

\newcommand{\mbW}{\mathbf{W}}

\newcommand{\mbA}{\mathbf{A}}
\newcommand{\mbK}{\mathbf{K}}
\newcommand{\mbB}{\mathbf{B}}
\newcommand{\mbU}{\mathbf{U}}
\newcommand{\mbM}{\mathbf{M}}

\DeclareMathOperator*{\argmax}{arg\,max} 
\begin{document}
	
\title{VAE-KRnet and its applications to variational Bayes}
\author{Xiaoliang Wan and Shuangqing Wei} 
\curraddr[X. Wan]{Department of Mathematics, 
and Center for Computation and Technology, 
Louisiana State University\\
Baton Rouge, LA 70803}
\email[X. Wan]{xlwan@lsu.edu}
\curraddr[S. Wei]{Division of Electrical \& Computer Engineering, 
Louisiana State University\\
Baton Rouge, LA 70803}
\email[S. Wei]{swei@lsu.edu}

\begin{abstract}
In this work, we have proposed a generative model, called VAE-KRnet, for density estimation or approximation, which combines the canonical variational autoencoder (VAE) with our recently developed flow-based generative model, called KRnet. VAE is used as a dimension reduction technique to capture the latent space, and KRnet is used to model the distribution of the latent variable. Using a linear model between the data and the latent variable, we show that VAE-KRnet can be more effective and robust than the canonical VAE. VAE-KRnet can be used as a density model to approximate either data distribution or an arbitrary probability density function (PDF) known up to a constant. VAE-KRnet is flexible in terms of dimensionality.  When the number of dimensions is relatively small, KRnet can effectively approximate the distribution in terms of the original random variable. For high-dimensional cases, we may use VAE-KRnet to incorporate dimension reduction. One important application of VAE-KRnet is the variational Bayes for the approximation of the posterior distribution. The variational Bayes approaches are usually based on the minimization of the Kullback-Leibler (KL) divergence between the model and the posterior. For high-dimensional distributions, it is very challenging to construct an accurate density model due to the curse of dimensionality, where extra assumptions are often introduced for efficiency. For instance, the classical mean-field approach assumes mutual independence between dimensions, {which often yields an underestimated variance due to  oversimplification. To alleviate this issue, we include into the loss the  maximization of the mutual information between the latent random variable and the original random variable, which helps keep more information from the region of low density such that the estimation of variance is improved.}  Numerical experiments have been presented to demonstrate the effectiveness of our model.
\end{abstract}

\maketitle

\section{Introduction}
The density estimation of high-dimensional data plays an important role in unsupervised learning, which is challenging due to the curse of dimensionality \cite{Scott2015}. In the last decade, deep generative modeling has made a lot of progress by incorporating with deep neural networks. Deep generative models are usually with likelihood-based methods, such as the autoregressive models \cite{Graves_2013,Oord_2016a,Oord_2016b,Papamakarios_2018}, variational autoencoders (VAE) \cite{Kingma_2014,Kingma_2016,HigginsMPBGBML17}, and flow-based generative models \cite{Dinh_2014,Rezende_2015,Dinh_2016,Dhariwal_2018,Zhang_2018,Berg_2019,Chen_2019,Dupont_2019}. One flexible model that does not need the likelihood is the generative adversarial network (GAN) \cite{Goodfellow_2014,Arjovsky_2017}, which seeks a Nash equilibrium of a zero-sum game between the generator and the discriminator. Recently, the coupling of different modeling strategies has also been explored. The flow-based model was coupled with GAN in \cite{Grover_2018} to obtain a likelihood for GAN;  The VAE, flow-based model and GAN were coupled in \cite{Zhu_2019} for more flexibility and efficiency. The main goal of deep generative models is to generate new data that are consistent with the underlying distribution of the available data. To achieve this, a specific density model is not a necessity, e.g., GAN manages to focus on the mapping from a standard Gaussian to the desired data distribution without using the likelihood. Other than GANs, deep generative models usually provide a density model, e.g., the flow-based models actually define an invertible transport map between two random variables which yields an explicit push-forward measure. A common characteristic of deep generative models is that they employ neural networks to model the mapping between high-dimensional inputs and outputs whenever needed. Such a strategy is proved to be very effective for application problems although the models are usually not easy to analyze due to the strong nonlinearity induced by neural networks.  

Classical density estimation techniques such as kernel density estimation and mixture of Gaussians, suffer severely from the curse of dimensionality, meaning that they are only effective for low-dimensional data. However, the approximation of high-dimensional distributions is often expected to alleviate the computational cost of sampling a complicated mathematical model. For example, a typical Uncertainty Quantification (UQ) model is a partial differential equation (PDE) subject to uncertainty. When studying  rare events in such a system, we must have an effective strategy to reduce the number of samples since  each sample corresponds to solving a PDE. One strategy is to use the reduced-order model to obtain the samples of the desired rare events followed by a density estimation step. The estimated distribution can then be coupled with the importance sampling technique for variance reduction \cite{Rubinstein2004,Giles_AN15,Wan_JCP20}. Another important example is the variational Bayes \cite{Blei_2018}. Sampling strategies such as Markov Chain Monte Carlo (MCMC) become less effective as the number of dimensions increases. The variational Bayes approach, which seeks the optimal approximation of the distribution in a family of density models, may be more effective for high-dimensional problems than sampling strategies. 

The available deep generative models usually focus on capturing the main features of the data instead of an accurate estimation of the density for that the dimensionality of the target data is often extremely high, e.g., high-resolution images that have millions of pixels. We are more interested in whether the strategies developed for deep generative models can be adapted as a density estimation technique with mathematical convergence. In \cite{Wan_KRnet}, we coupled the real NVP \cite{Dinh_2016} and the Knothe-Rosenblatt (KR) rearrangement to generate an invertible transport map, called KRnet, between the standard Gaussian and an arbitrary distribution.  In numerical experiments, KRnet has demonstrated a much better algebraic convergence than the original real NVP with respect to the number of model parameters. The drawback of constructing a transport map is that the dimensionality needs to be kept unchanged, which limits KRnet to a relatively small number of dimensions. In this work, we intend to couple KRnet and variational autoencoder (VAE) to obtain a more general model called VAE-KRnet. The basic idea is to use KRnet to model the prior distribution of the latent random variable identified by VAE and generalize  the underlying distribution of the encoder from Gaussian to an arbitrary one. We show that VAE-KRnet is more flexible than the canonical VAE by   
examining a linear model between the latent space and the data space. We then apply VAE-KRnet to the variational Bayes to approximate the posterior distribution. By varying the number of dimensions of the late space from zero (KRnet) to $d$ (VAE-KRnet), a wide range of data dimensions can be covered especially when the problem admits a significant dimension reduction. One common problem in variational Bayes is the possible underestimation of variance because the minimization of the Kullback-Leibler divergence is more in favor of the first-order moments than the second-order moments, especially when the model capability is not strong enough. To alleviate this issue, we take into account the mutual information when searching the latent random variable. In the loss we balance the contribution of two terms: the maximization of the mutual information between the latent random variable and the original random variable, and the minimization of the KL divergence between the density model and the original distribution. The relative importance of these two terms will be adjusted by a weight parameter. {By maximizing the mutual information we may keep more information from the region of low density and improve the estimation of the variance.}

Our paper is organized as follows. We first present a brief description of KRnet in the next section. We discuss the coupling of VAE and KRnet in section \ref{sec:VAE-KRnet}, and apply VAE-KRnet to variational Bayes in section \ref{sec:variational_bayes}. In section \ref{sec:numerical}, we study numerically the performance of VAE-KRnet, followed by a summary section.

\section{KRnet - An invertible transport map}
Let $\mu_{\bY}$ and $\mu_{\bZ}$ indicate the probability measures of random variables $\bY,\bZ\in\mathbb{R}^n$ respectively. A transport map ${T}:\bZ\rightarrow\bY$ is defined as ${T}_\#\mu_{\bZ}=\mu_{\bY}$, where ${T}_{\#}\mu_{\bZ}$ is the push-forward of the law $\mu_{\bZ}$ of $\bZ$ such that $\mu_{\bY}(B)=\mu_{\bZ}({T}^{-1}(B))$ for every Borel set $B$ \cite{Filippo2010}. The Knothe-Rosenblatt (K-R) rearrangement says that ${T}$ may have a lower-triangular structure such that 
\begin{equation}\label{eqn:KR}
\bz={T}^{-1}(\by)=f(\by)=\left[
\begin{array}{l}
f_1(y_1)\\
f_2(y_1,y_2)\\
\vdots\\
f_n(y_1,y_2,\ldots,y_n)
\end{array}
\right],
\end{equation}  
which can be regarded as a limit of a sequence of optimal transport maps when the quadratic cost degenerates \cite{Carlier_2010}. Combining the {triangular structure of the} K-R rearrangement and the technique real NVP \cite{Dinh_2016}, we have proposed an invertible mapping $\bz=\hat{f}(\by)=\hat{T}^{-1}(\by)$ such that $\hat{T}_{\#}\mu_{\bZ}$ can be used as a model for density estimation when data are provided for $\bY$ and a prior distribution is prescribed for $\bZ$ \cite{Spantini_2017,Wan_KRnet}. The invertible transport map $\hat{f}(\cdot)$ is called KRnet. In reality, we may consider a block-triangular version of the K-R rearrangement for more flexibility. Consider a partition of $\by=(\by_1,\ldots,\by_K)$, where $\by_i=(y_{i,1},\ldots,y_{i,m})$, where $1\leq K\leq n$ and $1\leq m\leq n$, and $\sum_{i=1}^K\textrm{dim}(\by_i)=n$. The transport map $\hat{f}(\by)$ then takes the following form:
\begin{equation}\label{eqn:KR-block}
\bz=\hat{f}(\by)=\left[
\begin{array}{l}
\hat{f}_1(\by_1)\\
\hat{f}_2(\by_1,\by_2)\\
\vdots\\
\hat{f}_K(\by_1,\ldots,\by_K)
\end{array}
\right].
\end{equation}  
Let $\mu_{\bZ}(d\bz)=p_{\bZ}(\bz)d\bz$, where $p_{\bZ}(\bz)$ is the probability density function (PDF). The KRnet $\hat{f}(\cdot)$ induces the following density model 
\begin{equation}\label{eqn:pdf_model}
p_{\bY}(\by)=p_{\bZ}(\hat{f}(\by))\left|\det\nabla_{\by} \hat{f}(\by)\right|,
\end{equation}
which can be easily sampled as $\bY=\hat{f}^{-1}(\bZ)$, thanks to the invertibility of $\hat{f}(\cdot)$.

\subsection{An overview of the layers in KRnet}\label{sec:layers_krnet}
The mathematical form of KRnet is an invertible composite mapping
\begin{equation}
\bz=f(\by)=f_{[m]}\circ f_{[m-1]}\circ\ldots\circ f_{[i]}\circ\ldots f_{[2]}\circ f_{[1]}(\by),
\end{equation} 
or
\begin{equation}
\by=f^{-1}(\bz)=f^{-1}_{[1]}\circ f^{-1}_{[2]}\circ\ldots\circ f^{-1}_{[i]}\circ\ldots f^{-1}_{[m-1]}\circ f^{-1}_{[m]}(\bz),
\end{equation}
where $f_{[i]}(\cdot)$ is a bijection that is often referred to as a layer. Simply speaking, KRnet modifies the data distribution of $\bY$ step by step through a large number of intermediate simple bijections to make it eventually consistent with a prescribed distribution of $\bZ$. We let $\by_{[0]}=\by$ indicate the initial state and $\by_{[i]}=f_{[i]}\circ\ldots\circ f_{[1]}(\by)$ an intermediate state. The main feature of KRnet is that the overall structure of the invertible mapping is lower (or upper) triangular. For the mapping from $\bY$ to $\bZ$, each dimension of $\bY$ will be deactivated at a certain stage and remain fixed until all dimensions become inactive. On the other hand, the inverse mapping from $\bZ$ to $\bY$ will activate the dimensions gradually. We now briefly introduce the layers used in KRnet, where each layer is a relatively simple bijection. More details about KRnet can be found in \cite{Wan_KRnet,Tang_ADDE}.

\emph{Squeezing layer} deactivates a certain number of components using a mask 
	\[
	{\bq} = (\underbrace{1,\ldots,1}_{k},\underbrace{0,\ldots,0}_{n-k}),
	\]
	which means that the components $\bq\odot\by_{[i]}$ will keep being updated and the rest components $(1-\bq)\odot\by_{[i]}$ will remain fixed from then on. Here  $\odot$ indicates the Hadamard product or component-wise product. So we deactivate the last $n-k$ components whenever needed.  
	
\emph{Rotation layer} provides a simple and trainable strategy to determine the dimensions that will be deactivated first. The rotation layer defines a rotation of the coordinate system through an orthonormal matrix for the current active dimensions:
	\[
	\hat{\by}_{[i]}=\hat{\mbW}\by_{[i]}=\left[
	\begin{array}{cc}
	\mbW&0\\
	0&\mbI
	\end{array}
	\right]\by_{[i]}=\left[
	\begin{array}{cc}
	\mathbf{L}&0\\
	0&\mbI
	\end{array}
	\right]
	\left[
	\begin{array}{cc}
	\mathbf{U}&0\\
	0&\mbI
	\end{array}
	\right]\by_{[i]},
	\]
	where $\mbW\in\mbbR^{k\times k}$ with $k$ being the number of 1's in ${\bq}$, and $\mbI\in\mbbR^{(n-k)\times(n-k)}$ is an identity matrix, and $\mbW=\mathbf{LU}$ is the LU decomposition of $\mbW$. The entries of $\mathbf{L}$ and $\mathbf{U}$ will be treated as trainable parameters except for the diagonal entries of $\mathbf{L}$ which are equal to 1. Intuitively we expect the rotation may put the most important dimensions at the beginning, which need further modifications. One implementation issue is that we usually optimize the trainable entries of $\mathbf{L}$ and $\mathbf{U}$ directly for simplicity without enforcing the orthonormality of $\mathbf{W}$, which works well in practice. 

\emph{Scale and bias layer} provides a simplification of batch normalization which is defined as  \cite{Szegedy_2015,Dhariwal_2018}
	\begin{equation}\label{eqn:actnorm}
	\hat{\by}_{[i]} = \ba\odot\by_{[i]}+\bb, 
	\end{equation}
	where $\ba$ and $\bb$ are trainable, and initialized by the mean and standard deviation of data. After the initialization, $\ba$ and $\bb$ will be treated as regular trainable parameters that are independent of the data. The scale and bias layer helps to improve the conditioning of deep net.

\emph{Affine coupling layer} is the most important layer for evolving the data. Consider a partition 
	$\by_{[i]}=(\by_{[i],1}^\mathsf{T},\by_{[i],2}^\mathsf{T})^\mathsf{T}$ with $\by_{[i],1}\in \mathbb{R}^m$ and $\by_{[i],2}\in\mathbb{R}^{n-m}$. The affine coupling layer is  defined as \cite{Wan_KRnet,Dinh_2016}
	\begin{align}
      \bz_1&=\by_{[i],1},\label{eqn:new_affine_1}\\
      \bz_2&=\by_{[i],2}\odot(1+\alpha\tanh(\bs(\by_{[i],1}))+e^{\bbeta}\odot\tanh(\bt(\by_{[i],1})),\label{eqn:new_affine_2}
	\end{align}
	where $\bs,\bt\in\mathbb{R}^{n-m}$ stand for scaling and translation functions depending only on $\by_{[i],1}$, {$0<\alpha<1$ is a hyperparameter and $\bbeta\in\mathbb{R}^n$ is trainable. We modified the original affine coupling layer in real NVP \cite{Dinh_2016} to improve the conditioning.}  Note that $\by_{[i],2}$ is updated linearly while the mappings $\bs(\by_{[i],1})$ and $\bt(\by_{[i],1})$ can be arbitrarily complicated, which are modeled as a neural network (NN),
	\begin{equation}\label{eqn:NN}
	(\bs, \bt) = \textsf{NN}(\by_{[i],1}).
	\end{equation}
	Then the Jacobi matrix is lower-triangular, and an inverse can be easily computed. The two parts of $\by_{[i]}$ will be updated alternately by a sequence of affine coupling layers, e.g., at the next affine coupling layer, the first part will be modified while the second part remains fixed.

\emph{Nonlinear invertible layer} defines a component-wise one-dimensional nonlinear mapping to map $\mathbb{R}$ to itself. {We decompose $\mathbb{R}=(-\infty,-a)\cup[-a,a]\cup(a,\infty)$ for $0<a<\infty$. The intervals $(-\infty,-a)\cup(a,\infty)$ and $[-a,a]$ will be mapped to themselves.  For $(-\infty,-a)\cup(a,\infty)$, a linear mapping is used, and for $[-a,a]$, a piecewise quadratic mapping is defined.} More specifically, we define
	\[
	z=\hat{F}(y)=\left\{
\begin{array}{rl}
\beta (y-a)+a,&y\in(-\infty,-a)\\
\phi^{-1}\circ F\circ\phi(y),&y\in[-a,a]\\
\beta (y+a)-a,&y\in(a,\infty),
\end{array}
\right.
	\] 
	where $\phi:[-a,a]\rightarrow[0,1]$ is an affine mapping, $\beta>0$ is a scaling factor, and 
	\begin{equation}
	F(x)=\int_0^xp(x)dx,\quad \forall x\in[0,1],
	\end{equation}
	where $p(x)$ can be regarded a PDF and $F(x)$ a cumulative distribution function. In particular, $p(x)$ will be defined as a piecewise linear function on a mesh of $[0,1]$ such that $F(x)$ is a quadratic function whose inverse can be computed explicitly. {Nonlinear invertible layer provides a component-wise change of variable for a standard Gaussian prior distribution.}

\subsection{Main structure of KRnet}
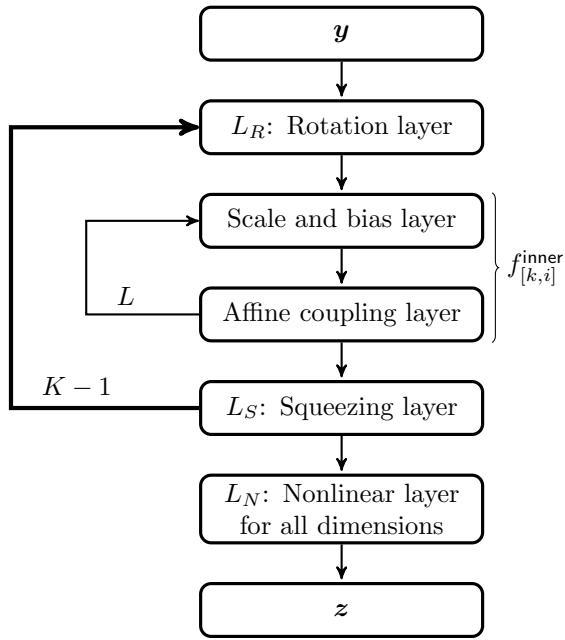
\begin{figure}
	\centering
	\begin{tikzpicture}
	[node distance=.5cm,
	start chain=going below,]
	\node[punktchain, join] (input) {$\by$};
	\node[punktchain, join] (rotation)      {$L_R$: Rotation layer};
	\node[punktchain, join] (scale)      {Scale and bias layer};
	\node[punktchain, join] (affine) {Affine coupling layer};
	\node[punktchain, join, ] (squeeze) {$L_S$: Squeezing layer};
	\node[punktchain, join, ] (nonlinear) {$L_N$: Nonlinear layer for all dimensions};
	\node[punktchain, join, ] (output) {$\bz$};
	\draw[tuborg, decoration={brace}] let \p1=(scale.north), \p2=(affine.south) in
	($(2, \y1)$) -- ($(2, \y2)$) node[tubnode] {$f_{[k,i]}^{\textsf{inner}}$};
	\draw[thick, ->]  
	(affine.west)
	-- ++ (-1.5cm, 0) 
	node[pos=0.65,above,sloped] {$L$} |-  (scale.west);
	\draw[ultra thick, ->]  
	(squeeze.west)
	-- ++ (-2.5cm, 0) 
	node[pos=0.65,above,sloped] {$K-1$} |-  (rotation.west);  
	\end{tikzpicture}
	\caption{The flow chart of KRnet.}
	\label{fig:structure_diagram}
\end{figure}

We are now ready to present the main structure of KRnet, which is illustrated in Figure \ref{fig:structure_diagram}. 
KRnet is mainly defined by two loops: outer loop $f_{[k]}^{\mathsf{outer}}(\cdot)$ and inner loop $f_{[k,i]}^{\mathsf{inner}}(\cdot)$ for a fixed $k$, where the outer loop has $K-1$ stages, corresponding to the $K-1$ mappings $f_i$ in equation \eqref{eqn:KR-block} with $i=2,\ldots,K$, and the inner loop has $L$ stages, indicating the length of a   chain of general coupling layers, see figure \ref{fig:structure_diagram}.
\begin{itemize}
	\item Outer loop. Let $f^{\mathsf{outer}}_{[k]}$ indicate one iteration of the outer loop. We have 
	\begin{equation}
	\bz=f(\by)=L_N\circ f^{\mathsf{outer}}_{[K-1]}\circ\ldots\circ f^{\mathsf{outer}}_{[1]}(\by).
	\end{equation}
	Let $\by_{[k]}=f^{\mathsf{outer}}_{[k]}(\by_{[k-1]})$ with $\by_{[0]}=\by$, and $i=1,\ldots,K-1$. Each $\by_{[k]}=(\by_{[k],1},\ldots,\by_{[k],K})$ has the same partition. The $i$th partition will remain unchanged after stage $K-i+1$. For example, $\by_{[k],K}$ will be updated only when $k=1$ and and $\by_{[k],K-1}$ will be fixed when $k>2$. This way, the number of effective dimensions decreases as $k$ increases. Once the outer loop is completed, the only active dimensions in $\by_{[K-1]}=(\by_{[K-1],1},\ldots,\by_{[K-1],K})$ will be $\by_{[K-1],1}$. We then apply the nonlinear invertible layer to all dimensions before the final output. 
	\item Inner loop. The inner loop mainly consists of a sequence of general coupling layers $f^{\mathsf{inner}}_{[k,i]}$, which includes one scale and bias layer and one affine coupling layer. We have 
	\begin{equation}
	f^{\mathsf{outer}}_{[k]}=L_S\circ f^{\mathsf{inner}}_{[k,L]}\circ\ldots\circ f^{\mathsf{inner}}_{[k,1]}\circ L_R,
	\end{equation}
	where $L_R$ is a rotation layer, and $L_S$ is a squeezing layer. 
\end{itemize}
The main effectiveness of KRnet comes from the depth determined by both $K$ and $L$. The rotation layers and the nonlinear invertible layers may be switched off initially to reduce the model complexity. Once the KRnet is trained, these layers can be switched on for further refinement.  

\section{Coupling VAE and KRnet}\label{sec:VAE-KRnet}
Due to the invertibility, KRnet maps a variable to another variable of the same dimension. To enhance its capability to deal with high-dimensional data, we will integrate it into the framework of variational autoencoder (VAE), which provides a dimension reduction technique for density estimation. We employ VAE to identify the latent space and KRnet to generalize the modeling for  the prior distribution and the encoder. 

\subsection{Variational autoencoder (VAE)}
We briefly recall the variational autoencoder \cite{Kingma_2014}. Assume that there exists a latent random variable $\bX\in\mathbb{R}^d$ with $d\ll n$ with a marginal distribution $p_{\bX,\btheta}(\bx)$, where $\btheta$ includes the model parameters. The joint distribution $p_{\bX,\bY,\btheta}$ of $\bX$ and $\bY$ is then described by the conditional distribution $p_{\bY|\bX,\btheta}(\by|\bx)$, i.e., $p_{\bX,\bY,\btheta}=p_{\bY|\bX,\btheta}p_{\bX,\btheta}$. 

The target is to approximate the posterior distribution $p_{\bX|\bY^{(i)},\btheta}(\bx|\by^{(i)})$ which will be modeled by a family of parameterized PDFs $q_{\bX|\bY^{(i)},\bphi}(\bx|\by^{(i)})$. Here we add a superscript $*^{(i)}$ to emphasize that the random variable $\bY^{(i)}$ corresponds to one sample in the training set. The optimal parameters $\btheta$ and $\bphi$ are determined by minimizing the KL divergence
\begin{align}
&D_{\mathsf{KL}}(q_{\bX|\bY^{(i)},\bphi}\|p_{\bX|\bY^{(i)},\btheta})\nonumber\\
=&D_{\mathsf{KL}}(q_{\bX|\bY^{(i)},\bphi}\|p_{\bX,\theta})-\int q_{\bX|\bY^{(i)},\bphi}\log p_{\bY^{(i)}|\bX,\theta}d\bx+\log p_{\bY^{(i)}}\geq0.\label{eqn:q_to_p}
\end{align}
The minimization of $D_{\mathsf{KL}}(q_{\bX|\bY^{(i)},\bphi}\|p_{\bX|\bY^{(i)},\btheta})$ is equivalent to the maximization of the variational lower bound of $\log p_{\bY^{(i)}}$, which is defined as 
\begin{equation}\label{eqn:var_lower_bound}
\mathcal{L}_{\btheta,\bphi}(\by^{(i)})=-D_{\mathsf{KL}}(q_{\bX|\bY^{(i)},\bphi}\|p_{\bX,\theta})+\int q_{\bX|\bY^{(i)},\bphi}\log p_{\bY^{(i)}|\bX,\theta}d\bx.
\end{equation}
If there are $N$ samples in the training set, the variational lower bound of the log-likelihood $\log p_{\mathbf{Y}}$ is 
\begin{equation}
\mathcal{L}_{\btheta,\bphi}(\mathbf{y})=\sum_{i=1}^N\mathcal{L}_{\btheta,\bphi}(\by^{(i)}),
\end{equation}
where $\mathbf{y}$ includes all the data $\{\by^{(i)}\}_{i=1}^N$ in the training set. 

To apply VAE we need to specify three PDF models respectively for $p_{\bY^{(i)}|\bX,\btheta}$, $q_{\bX|\bY^{(i)},\bphi}$ and $p_{\bX,\btheta}(\bx)$. In the canonical VAE, the following models are used:
\begin{align}
p_{\bY^{(i)}|\bX,\btheta}&=\mathcal{N}(\bmu_{\mathsf{de},\btheta}(\bx),\mathrm{diag}(\bsigma_{\mathsf{de},\btheta}^{\odot2}(\bx))),\label{eqn:vae_decoder}\\
q_{\bX|\bY^{(i)},\bphi}&=\mathcal{N}(\bmu_{\mathsf{en},\bphi}(\by^{(i)}),\mathrm{diag}(\bsigma_{\mathsf{en},\bphi}^{\odot2}(\by^{(i)}))),\label{eqn:vae_encoder}\\
p_{\bX,\btheta}&=\mathcal{N}(0,\mathbf{I}),\label{eqn:vae_prior}
\end{align}
where $*^{\odot 2}$ means that the square operation is component-wise. 
From the viewpoint of dimension reduction, it is often a good choice to assume that $p_{\bY^{(i)}|\bX,\btheta}$ is Gaussian with independent components. The posterior distribution $p_{\bX|\bY^{(i)},\btheta}$ is in general intractable, and an approximation model $q_{\bX|\bY^{(i)},\bphi}$ is needed, which is also chosen as a multivariate Gaussian with independent components. Then $(\bmu_{\mathsf{en},\bphi}(\by),\bsigma_{\mathsf{en},\bphi}(\by))$ serves as the encoder and $(\bmu_{\mathsf{de},\btheta}(\bx),\bsigma_{\mathsf{de},\btheta}(\bx))$ serves as the decoder. The prior distribution $p_{\bX,\btheta}$ is simply assumed to be a standard Gaussian $\mathcal{N}(0,\mathbf{I})$. Furthermore, both encoder and decoder are modeled by neural networks. 

\subsection{VAE for a linear model}\label{sec:VAE_linear}
{VAE has three components: the prior, the encoder and the decoder, each of which is either a simple Gaussian or a diagonal Gaussian. Let us first examine the relation of these three components in terms of a linear model for dimension reduction}
\begin{equation}\label{eqn:linear_reduction}
\bY=\mbA\bX+\bxi,
\end{equation}
where $\bxi\in\mathbb{R}^n$, $\bxi\sim\mathcal{N}(0,\sigma^2\mathbf{I})$, $\mbA\in\mathbb{R}^{n\times d}$, and $\bxi$ is independent of $\bX$. We regard $\bX$ as a latent random variable with $d\ll n$. The joint distribution of $\bX$ and $\bY$ is 
\begin{equation}
p_{\bX,\bY}=p_{\bX}\cdot\mathcal{N}(\mbA\bx,\sigma^2\mathbf{I})
\end{equation}
After the decoder $p_{\bY|\bX}=\mathcal{N}(\mbA\bx,\sigma^2\mathbf{I})$ is specified, we look at the relation between $p_{\bX}$ and $q_{\bY|\bX}$: 
\begin{lemma}
Let $p_{\bX}=\mathcal{N}(0,\mathbf{\Sigma}_{\bX})$ in model \eqref{eqn:linear_reduction} with $\mathbf{\Sigma}_{\bX}$ being positive definite. From the modeling point of view, the encoder $q_{\bX|\bY}=\mathcal{N}(\bmu_{\mathsf{en}}(\by),\bsigma_{\mathsf{en}}^{\odot2}(\by))$ of the canonical VAE is able to exactly recover the true posterior distribution $p_{\bX|\bY}$. 
\end{lemma}
\begin{proof}
The prior is chosen as $p_{\bZ}=\mathcal{N}(0,\mathbf{I})$ in VAE, implying that we may consider a transformed model 
\[
\bY=\mbA(\mbB\mbU\bZ)+\bxi,
\] 
where $\bX=\mbB\mbU\bZ$ with $\mbB,\mbU\in\mathbb{R}^{d\times d}$ and $\mbU$ being a unitary matrix. The reason that we include the unitary matrix $\mbU$ will be clear later. It is easy to see that the variable $\bZ|_{\by}$ is Gaussian subject to a covariance matrix
\begin{equation}\label{eqn:Sigma_inv}
\mathbf{\Sigma}_{\bZ|\by}=(\mathbf{I}+\sigma^{-2}\mbU^{\mathsf{T}}\mbB^{\mathsf{T}}\mbA^{\mathsf{T}}\mbA\mbB\mbU)^{-1}.
\end{equation}
If the encoder of VAE is able to exactly recover $\mathbf{\Sigma}_{\bZ|\by}$, we need the existence of a linear mapping $\mbB\mbU$ such that $\mathbf{\Sigma}_{\bZ|\by}$ is diagonal. Note that
\[
\mathrm{Cov}(\bX)=\mathbf{\Sigma}_{\bX}=\mbB\mbU\mathrm{Cov}(\bZ)\mbU^{\mathsf{T}}\mbB^{\mathsf{T}}=\mbB\mbB^{\mathsf{T}},
\]
which means that we can let $\mbB=\mathbf{\Sigma}_{\bX}^{1/2}$. According to the spectral theorem of symmetric matrices, we know there exists a unitary matrix $\mbU$ such that $\mbU^\mathsf{T}\mbB^{\mathsf{T}}\mbA^{\mathsf{T}}\mbA\mbB\mbU$ is diagonal. This concludes the lemma. \qed
\end{proof}

\begin{remark}\label{rmk:encoder}
For the linear model \eqref{eqn:linear_reduction} with a Gaussian prior, the canonical VAE is able to model the posterior, where the encoder needs to take care of three mappings: (1) the mapping $\mbB$, and (2) the rotation $\mbU$, and (3) the matrix operations in equation \eqref{eqn:Sigma_inv}. The mapping $\mbB$ transfers a general Gaussian to a standard one, and the rotation $\mbU$ makes the covariance matrix of $\bZ|_{\by}$ diagonal. Under the assumption that the diagonal decoder provides a reasonable model for dimension reduction, we know from this case study that the encoder needs to balance the following two issues:
\begin{enumerate}
\item Map the prior distribution $p_{\bX}$ to a standard Gaussian $\mathcal{N}(0,\mbI)$;  
\item Map the posterior distribution $p_{\bX|\bY}$ to a diagonal Gaussian. 
\end{enumerate}
So the effectiveness of the canonical VAE depends on how well these two issues can coexist, which is obviously problem dependent.  
\end{remark}

\subsection{Generalize the prior}
Assume that $\bX\sim p_G=\mathcal{N}(0,\mathbf{I})$ following the canonical VAE. We introduce another random variable $\bZ$ satisfying $\bX=f_{\mathsf{pr},\bbeta}(\bZ)$, where $f_{\mathsf{pr},\bbeta}(\cdot)$ is a nonlinear bijection with $\bbeta$ being the model parameter, e.g., KRnet. We have $p_{\bZ,\bbeta}(\bz)=p_{G}(f_{\mathsf{pr},\bbeta}(\bz))|\nabla_{\bz}f_{\mathsf{pr},\bbeta}(\bz)|$. We now compare the two cases, where the latent spaces are defined by $\bX$ and $\bZ$ respectively. We also assume that $q_{\bX|\bY,\bphi}$ and $q_{\bZ|\bY,\bphi}$ are defined by the same model, i.e., Gaussian, and so are $p_{\bY|\bX,\btheta}$ and $p_{\bY|\bZ,\btheta}$. In other words, only the model for the prior is changed. 
\begin{lemma}
{If $f_{\mathsf{pr},\bbeta}(\cdot)$ induces a density model that is able to approximate any $d$-dimensional PDF arbitrarily well,} then there exists a parameter $\bbeta=\tilde{\bbeta}$ such that VAE can reach a larger variational lower bound in terms of the random variable $\bZ$ rather than $\bX$. 
\end{lemma}
\begin{proof}
Noting that 
\begin{equation}
\lim_{N\rightarrow\infty}\frac{1}{N}\mathcal{L}^{\bX}_{\bphi,\btheta}(\by)=\lim_{N\rightarrow\infty}\sum_{i=1}^N\frac{1}{N}\mathcal{L}^{\bX}_{\bphi,\btheta}(\by^{(i)})=
\mathbb{E}_{p_{\bY}}\mathcal{L}^{\bX}_{\bphi,\btheta}(\bY),
\end{equation}
where the superscript $\bX$ indicates that the latent space is defined by $\bX\sim\mathcal{N}(0,\mathbf{I})$ and 
\begin{equation}\label{eqn:elbo_cnt}
\mathbb{E}_{p_{\bY}}\left[\mathcal{L}^{\bX}_{\bphi,\btheta}(\bY)\right]
=-D_{\mathsf{KL}}(q_{\bX|\bY,\bphi}p_{\bY}\|p_{\bX,\btheta}p_{\bY})+
\mathbb{E}_{p_{\bY}q_{\bX|\bY,\bphi}}
\left[\log p_{\bY|\bX,\btheta}\right].
\end{equation}
For simplicity, we consider $\mathbb{E}_{p_{\bY}}\left[\mathcal{L}^{\bX}_{\bphi^*,\btheta^*}(\bY)\right]$ instead of the average of $\mathcal{L}^{\bX}_{\bphi,\btheta}(\by^{(i)})$, where 
\begin{equation}
(\bphi^*,\btheta^*)=\argmax_{\bphi,\btheta}\mathbb{E}_{p_{\bY}}\left[\mathcal{L}^{\bX}_{\bphi,\btheta}(\bY)\right].
\end{equation}
By definition, we have
\begin{align*}
&\mathbb{E}_{p_{\bY}}\left[\mathcal{L}^{\bX}_{\bphi^*,\btheta^*}(\bY)\right]-\mathbb{E}_{p_{\bY}}\left[\mathcal{L}^{\bZ}_{\bphi^*,\btheta^*,\bbeta}(\bY)\right]\\
&=D_{\mathsf{KL}}(q_{\bZ|\bY,\bphi^*}p_{\bY}\|p_{\bZ,\bbeta}p_{\bY})
-D_{\mathsf{KL}}(q_{\bX|\bY,\bphi^*}p_{\bY}\|p_{G}p_{\bY}),
\end{align*}
since only the prior depends on $\bbeta$ and the encoders and decoders are the same for both $\bX$ and $\bZ$.
We have
\begin{align*}
&D_{\mathsf{KL}}(q_{\bZ|\bY,\bphi^*}p_{\bY}\|p_{\bZ,\bbeta}p_{\bY})
-D_{\mathsf{KL}}(q_{\bX|\bY,\bphi^*}p_{\bY}\|p_{G}p_{\bY})\\
=&-\int q_{\bZ|\bY,\bphi^*}p_{\bY}\log p_{\bZ,\bbeta} d\bz d\by+\int q_{\bX|\bY,\bphi^*}p_{\bY}\log p_{G} d\bx d\by\\
=&-\int q_{\bZ,\bphi^*}\log p_{\bZ,\bbeta} d\bz+\int q_{\bX,\bphi^*}\log p_G d\bx\\
=&D_{\mathsf{KL}}(q_{\bZ,\bphi^*}\|p_{\bZ,\bbeta})-D_{\mathsf{KL}}(q_{\bX,\bphi^*}\|p_{G}),
\end{align*}
where $q_{\bZ,\bphi^*}$ and $q_{\bX,\bphi^*}$ refer to the same marginal distribution because by assumption the joint distributions $q_{\bZ|\bY,\bphi^*}p_{\bY}$ and $q_{\bX|\bY,\bphi^*}p_{\bY}$ are the same. So if $q_{\bX,\bphi^*}\neq p_G$, there always exists $\tilde{\bbeta}$ such that 
\[
D_{\mathsf{KL}}(q_{\bZ,\bphi^*}\|p_{\bZ,\tilde{\bbeta}})< D_{\mathsf{KL}}(q_{\bX,\bphi^*}\|p_{G})
\]
{since by assumption the density model induced by $\bX=f_{\mathsf{pr},\bbeta}(\bZ)$ can approximate the PDF $q_{\bZ,\bphi^*}$ arbitrarily well}. We then have
\[
\mathbb{E}_{p_{\bY}}\left[\mathcal{L}^{\bZ}_{\bphi^*,\btheta^*,\tilde{\bbeta}}(\bY)\right]>\mathbb{E}_{p_{\bY}}\left[\mathcal{L}^{\bX}_{\bphi^*,\btheta^*}(\bY)\right],
\]
implying that $\bZ$ provides a better latent space than $\bX$. \qed
\end{proof}

\subsection{Generalize the encoder}
We now look at the encoder $q_{\bX|\bY}$. Note 
\begin{equation}
p_{\bY}p_{\bX|\bY}=p_{\bX,\bY}=p_{\bX}p_{\bY|\bX},
\end{equation}
where $p_{\bX,\bY}$ is the joint distribution. In this equation, three PDFs, i.e., $p_{\bY|\bX}$, $p_{\bX|\bY}$ and $p_{\bX}$, will be modeled or approximated by Gaussians in the canonical VAE as shown in equations \eqref{eqn:vae_decoder}-\eqref{eqn:vae_prior}.  Although it is straightforward to assume that $p_{\bY|\bX}$ is a diagonal Gaussian from the viewpoint of model reduction, modeling  $p_{\bX|\bY}$ as a diagonal Gaussian is mainly for tractability. Let us consider the following optimization problem
\begin{equation}
(\bmu^*,\mathbf{\Sigma}^*)=\min_{(\bmu,\mathbf{\Sigma})}D_{\mathsf{KL}}[p_{\bX|\bY}\|\mathcal{N}(\bmu,\mathbf{\Sigma})],
\end{equation}
which yields the optimal Gaussian that approximates $p_{\bX|\bY}$.  More specifically, we have
\begin{equation}\label{eqn:opt_mu}
\bmu^*=\int p_{\bX|\bY}\bx d\bx=\mathbb{E}_{p_{\bX|\bY}}[\bX].
\end{equation}
and
\begin{equation}\label{eqn:opt_Sigma}
\mathbf{\Sigma}^*=\int p_{\bX|\bY}(\bx-\bmu)(\bx-\bmu)^\mathsf{T}d\bx=\mathbb{E}_{p_{\bX|\bY}}\left[(\bx-\bmu)(\bx-\bmu)^\mathsf{T}\right].
\end{equation}
Equation \eqref{eqn:opt_Sigma} shows that to approximate $p_{\bX|\bY}$ with a Gaussian, the encoder \eqref{eqn:vae_encoder} with a diagonal covariance matrix is in general not enough.

A straightforward way to generalize the encoder of the canonical VAE is to consider a full covariance matrix, i.e., $\mathcal{N}(\bmu_{\mathsf{en},\bphi}(\by),\mathbf{\Sigma}_{\mathsf{en},\bphi}(\by))$.  We here propose a simpler strategy. We let 
\begin{equation}
f_{\mathsf{en},\balpha}\left(\frac{\bX|{\by}-\bmu_{\mathsf{en},\bphi}(\by)}{\bsigma_{\mathsf{en},\bphi}(\by)}\right)\sim \mathcal{N}(0,\mathbf{I}),
\end{equation}
where $f_{\mathsf{en,\balpha}}$ is a KRnet. 
In other words, we can write 
\begin{equation}\label{eqn:x_y_xi}
\bX|{\by}=\bmu_{\mathsf{en},\bphi}(\by) + \bsigma_{\mathsf{en},\bphi}(\by)\odot f^{-1}_{\mathsf{en},\balpha}(\bxi),
\end{equation}
where $\bxi\sim\mathcal{N}(0,\mathbf{I})$. If we let $f_{\mathsf{en},\balpha}(\cdot)$ be an identity mapping, the original encoder \eqref{eqn:vae_encoder} is recovered. Let us look at the linear model \eqref{eqn:linear_reduction} again. If the prior $p_{\bX}=\mathcal{N}(\bmu_{\mathsf{pr}},\mathbf{\Sigma}_{\mathsf{pr}})$ is an arbitrary Gaussian, it can be obtained that the covariance matrix for $\bX|\by$ is 
\[
\mathbf{\Sigma}=(\mathbf{\Sigma}^{-1}_{\mathsf{pr}}+\sigma^{-2}\mbA^\mathsf{T}\mbA)^{-1}.
\]
Let $\bmu_{\mathsf{en},\bphi}(\by)=\mathbb{E}_{p_{\bX|\bY}}[\bX]$, $\bsigma_{\mathsf{en},\bphi}(\by)=\boldsymbol{1}$ and $f^{-1}_{\mathsf{en},\balpha}(\bxi)=\mathbf{\Sigma}^{1/2}\bxi$. The encoder \eqref{eqn:x_y_xi} is able to recover $p_{\bX|\bY}$ exactly for the linear model \eqref{eqn:linear_reduction} with any Gaussian priors, where the prior can be modeled directly by $f_{\mathsf{pr},\bbeta}(\cdot)$ and the correlation of $\bX|\by$ can be taken care of by $f^{-1}_{\mathsf{en},\balpha}(\cdot)$, in contrast to the canonical VAE, where the diagonal Gaussian encoder needs to achieve what both $f_{\mathsf{pr},\bbeta}(\cdot)$ and $f_{\mathsf{en},\balpha}^{-1}(\cdot)$ do. In this sense, the model becomes more flexible since both $f_{\mathsf{pr},\bbeta}(\cdot)$ and $f_{\mathsf{en},\balpha}^{-1}(\cdot)$ can help to maintain the diagonal form of the encoder.  

\subsection{VAE-KRnet}
\begin{figure}	
\center{
		\includegraphics[width=0.8\textwidth]{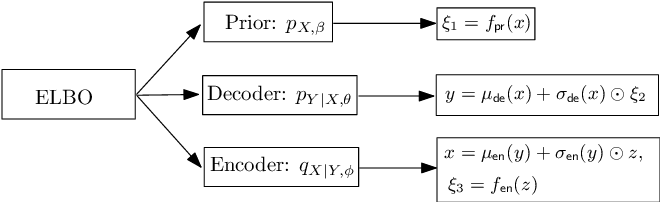}
	}
	\caption{Generalized variational autoencoder, where two flow-based generative models $f_{\mathsf{pr}}(\cdot)$ and $f_{\mathsf{en}}(\cdot)$ are integrated.}\label{fig:model_component}
\end{figure}

To this end, we have a simple strategy to couple VAE and KRnet. Within the framework of VAE, we keep the original decoder $\mathcal{N}(\bmu_{\mathsf{de},\btheta}(\bx),\mathrm{diag}(\bsigma_{\mathsf{de},\btheta}^{\odot2}(\bx)))$, but generalize the original prior $p_G$ and the encoder $\mathcal{N}(\bmu_{\mathsf{en},\bphi}(\by),\mathrm{diag}(\bsigma_{\mathsf{en},\bphi}^{\odot2}(\by)))$ by incorporating two KRnets $f_{\mathsf{pr},\bbeta}(\cdot)$ and $f_{\mathsf{en},\balpha}(\cdot)$, respectively, as demonstrated in figure \ref{fig:model_component}.

We now define the variational lower bound of VAE-KRnet using the following PDFs:
\begin{align}
p_{\bX,\bbeta}&=p_{G}(f_{\mathsf{pr},\bbeta}(\bx))|\nabla_{\bx}f_{\mathsf{pr},\bbeta}(\bx)|,\label{eqn:coupled_prior}\\
q_{\bX|\bY,\bphi,\balpha}&=p_{G}\left(f_{\mathsf{en},\balpha}\left(\frac{\bx-\bmu_{\mathsf{en},\bphi}(\by)}{\bsigma_{\mathsf{en},\bphi}(\by)}\right)\right)|\nabla_{\bx}f_{\mathsf{en},\balpha}(\bx)|,\label{eqn:coupled_xy}\\
p_{\bY|\bX,\btheta}&=\mathcal{N}(\bmu_{\mathsf{de},\btheta}(\bx),\mathrm{diag}(\bsigma_{\mathsf{de},\btheta}^{\odot2}(\bx)))\label{eqn:coupled_yx}.
\end{align}
The variational lower bound can be easily approximated by the Monte Carlo method. In particular, the so-called reparameterization trick \cite{Dinh_2014} can be employed. Using equation \eqref{eqn:x_y_xi}, the samples for the PDF $q_{\bX|\bY,\bphi,\balpha}(\bx|\by^{(i)})$ can be represented as  
\begin{equation}\label{eqn:x_y_sample}
\bx^{(i,j)}=\bmu_{\mathsf{en},\bphi}(\by^{(i)})+\bsigma_{\mathsf{en},\bphi}(\by^{(i)})\odot \bz^{(j)},\quad j=1,\ldots,J,
\end{equation}
where $\bz^{(j)}=f_{\mathsf{en},\balpha}^{-1}(\bxi^{(j)})$ and $\bxi\sim\mathcal{N}(0,\mathbf{I})$. 
The variational lower bound \eqref{eqn:var_lower_bound} will be approximated as
\begin{equation}
\mathcal{L}^{\bX}_{\btheta,\bphi,\bbeta,\balpha}(\by^{(i)})\approx\frac{1}{J}\sum_{j=1}^J\log\frac{ p_{\bY^{(i)}|\bX,\btheta}(\by^{(i)}|\bx^{(i,j)})p_{\bX,\bbeta}(\bx^{(i,j)})}{q_{\bX|\bY^{(i)},\bphi,\balpha}(\bx^{(i,j)}|\by^{(i)})}.
\end{equation} 
For simplicity, we can just let $J=1$ by noting that 
\begin{align}
\mathbb{E}_{p_{\bY}}\left[\mathcal{L}^{\bX}\right]&=\mathbb{E}_{p_{\bY}q_{\bX|\bY}}\left[\log\frac{p_{\bY|\bX}(\by|\bx)p_{\bX}}{q_{\bX|\bY}}\right]
\nonumber\\
&\approx\frac{1}{N}\sum_{i=1}^N\mathcal{L}^{\bX}_{\btheta,\bphi,\bbeta,\balpha}(\by^{(i)},\bx^{(i)})=
\frac{1}{N}\mathcal{L}^{\bX}_{\btheta,\bphi,\bbeta,\balpha}(\mathbf{y}),
\end{align}
where $(\by^{(i)},\bx^{(i)})$ corresponds to one sample from the joint PDF $p_{\bY}q_{\bX|\bY}$. 

\begin{remark}
It is seen that VAE-KRnet naturally integrates KRnet into the framework of VAE. Compared to other flow-based models, KRnet has its advantages. We here mainly comment on the invertibility of the transport map. Most flow-based generative models, which are able to provide an explicit likelihood, can be regarded as a transport map which requires either discrete or continuous invertibility. Discrete invertibility includes NICE \cite{Dinh_2014}, real NVP \cite{Dinh_2016}, planar flow \cite{Rezende_2015}, inverse autoregressive flow \cite{Kingma_2016}, Sylvester flow \cite{Berg_2019}, KRnet \cite{Wan_KRnet}, etc. Continuous invertibility mainly refers to neural ODE \cite{Chen_2019} and its variants subject to either augmentation \cite{Dupont_2019} or regularization \cite{Yang_2019,Finlay_2020}. The continuous invertibility can only be achieved in terms of the original ODE instead of the discretized ODE, which implies that the gradient of the loss may not be accurately computed through the adjoint method. Discrete invertibility such as the planar flow and the Sylvester flow depends on the inverse of an activation function, e.g., $\tanh$. The mapping from a finite interval to an infinite interval may have an issue on the robustness. To enhance VAE with flow-based generative models, we expect that the invertibility of the flow-based generative models can be easily maintained such that the overall model is relatively robust. The inverse autoregressive model and the real NVP, which is a generalization of NICE, can achieve discrete invertibility easily and exactly. However, the inverse autoregressive flow is more like a probabilistic model instead of a dynamical one, where the model structure is determined by the decomposition of the PDF in terms of conditional distributions. The real NVP defines a transport map, but its drawback is the information exchange among dimensions becomes less effective as the depth increases. By integrating the triangular structure of the Knothe-Rosenblatt rearrangement, KRnet, as a generalization of the real NVP, alleviates this issue such that the model capability has been significantly improved while the exact invertibility is kept.  
\end{remark}

\section{Density approximation}\label{sec:variational_bayes}
VAE-KRnet provides a family of probability density models that may be used for density estimation or approximation when 
data are available or the PDF is known up to a constant. If dimension reduction is not considered, we use  KRnet instead of VAE-KRnet. The configuration of KRnet is a trade-off between the number $K$ of outer-loop iterations and the number $L$ of inner-loop iterations. For a large $n$, we need to reduce at least one of the two numbers $K$ and $L$ for KRnet such that the model is affordable. Another way to deal with high dimensionality is to consider VAE-KRnet, where a low-dimensional latent space is sought. Since in reality problem solutions often admits a low-dimensional approximation, VAE-KRnet provides a mechanism to adapt the properties of the problem into density estimation. For simplicity, we use the subscript $*_{\btheta}$ to indicate a PDF model with a general model parameter $\btheta$ in this section. 

\subsection{Metrics for seeking the latent random variable}\label{sec:obj_v2}
The canonical VAE seeks the latent random variable based on available data by minimizing the KL divergence between $q_{\bX|\bY}$ and $p_{\bX|\bY}$ for any $\by$ (see equation \eqref{eqn:q_to_p}). Another way to achieve this is as follows:
\begin{lemma}
Maximizing $\mathbb{E}_{p_{\bY}}\left[\mathcal{L}^{\bX}\right]$ in VAE is equivalent to minimizing the following K-L divergence:
	\begin{equation}\label{eqn:VAE_v2}
	D_{\mathsf{KL}}(q_{\bX|\bY}p_{\bY}\|p_{\bY|\bX}p_{\bX}),
	\end{equation}
	which yields $q_{\bX|\bY}p_{\bY}=p_{\bY|\bX}p_{\bX}$ when the minimum is reached.
\end{lemma}
\begin{proof}
	 Equation \eqref{eqn:VAE_v2} can be rewritten as:
	\begin{align*}
	&D_{\mathsf{KL}}(q_{\bX|\bY}p_{\bY}\|p_{\bY|\bX}p_{\bX})\\
	=&\int q_{\bX|\bY}p_{\bY}\log\frac{q_{\bX|\bY}p_{\bY}}{p_{\bY|\bX}p_{\bX}}d\bx d\by\\
	=&\int q_{\bX|\bY}p_{\bY}\left(\log\frac{q_{\bX|\bY}p_{\bY}}{p_{\bX}p_{\bY}}+\log\frac{p_{\bY}}{p_{\bY|\bX}}\right)d\bx d\by\\
	=&-\mathbb{E}_{p_{\bY}}\left[\mathcal{L}^{\bX}\right]+\int q_{\bX|\bY}p_{\bY}\log p_{\bY}d\bx d\by\\
	=&-\mathbb{E}_{p_{\bY}}\left[\mathcal{L}^{\bX}\right]-h(\bY)\geq 0.
	\end{align*}
	So minimizing $D_{\mathsf{KL}}(q_{\bX|\bY}p_{\bY}\|p_{\bY|\bX}p_{\bX})$ is the same as maximizing $\mathbb{E}_{p_{\bY}}\left[\mathcal{L}^{\bX}\right]$, since the differential entropy $h(\bY)$ is not related to the models. \qed
\end{proof}
When we only have the samples of $p_{\bY}$, it is convenient to maximize the variational lower bound  $\mathbb{E}_{p_{\bY}}\left[\mathcal{L}^{\bX}\right]$. We also note that the relation $q_{\bX|\bY}p_{\bY}=p_{\bY|\bX}p_{\bX}$ can be obtained by minimizing 
	\begin{equation}\label{eqn:VAE_v3}
	D_{\mathsf{KL}}(p_{\bY|\bX}p_{\bX}\|q_{\bX|\bY}p_{\bY}),
	\end{equation}
where we switch the two distributions in equation \eqref{eqn:VAE_v2}. Due to the asymmetry of the K-L divergence, equations \eqref{eqn:VAE_v2} and \eqref{eqn:VAE_v3} are different from the computation point of view although they yield the same minimum and minimizer. {Equation \eqref{eqn:VAE_v3} is useful when $p_{\bY}$ is known up to a constant and data are not available. For this case the K-L divergence \eqref{eqn:VAE_v3} may be approximated by the samples generated in terms of $p_{\bX}$ and $p_{\bY|\bX}$.}

\subsection{When data are available}
For this case, we maximize the variational lower bound using the data of $\bY$. After obtaining the trained PDFs $p_{\bX,\btheta}$ and $p_{\bY|\bX,\btheta}$, we can approximate the marginal PDF of $\bY$:
\begin{equation}\label{eqn:p_Y_smaple}
p_{\bY,\btheta}(\by)=\mathbb{E}_{p_{\bX,\btheta}}[p_{\bY|\bX,\btheta}]\approx\frac{1}{N}\sum_{i=1}^Np_{\bY|\bX,\btheta}(\by|\bx^{(i)}),
\end{equation}
where the samples $\bx^{(i)}=f^{-1}_{\mathsf{pr},\bbeta}(\bxi^{(i)})$ with $\bxi\sim\mathcal{N}(\mathbf{0},\mathbf{I})$.  A more efficient  way to compute $p_{\bY,\btheta}(\by)$ is 
\begin{equation}\label{eqn:p_Y_IS}
p_{\bY,\btheta}(\by)=\mathbb{E}_{q_{\bX|\bY,\btheta}}\left[\frac{p_{\bY|\bX,\btheta}p_{\bX,\btheta}}{q_{\bX|\bY,\btheta}}\right]\approx\frac{1}{N}
\sum_{i=1}^N\frac{p_{\bY|\bX,\btheta}(\by|\bx^{(i)})p_{\bX,\btheta}(\bx^{(i)})}{q_{\bX|\bY,\btheta}(\bx^{(i)}|\by)}
\end{equation}
which uses the posterior to implement importance sampling. If the posterior is well approximated, the variance should be small such that less samples are needed compared to equation \eqref{eqn:p_Y_smaple}. 

\subsection{When PDF is available}
For many cases, we need to sample or approximate an arbitrary PDF known up to a constant, e.g., the posterior in Bayesian inference. {As an alternative of sampling methods such as MCMC, variational Bayes methods seek the best approximation of the posterior within a given family of density models. One classical variational Bayes approach is the mean-field approximation, where mutual independence is assumed for a partition of the latent random variable. In this work, we use VAE-KRnet as the variational distribution. We pay particular attention to one common problem of variational Bayes methods, which is the underestimation of variance. To alleviate this issue, we introduce the maximization of mutual information when seeking the latent random variable of VAE.} 

\subsubsection{KRnet}
Let $p_{\bY}=C^{-1}\hat{p}_{\bY}$ be a PDF, where $C$ is an unknown  normalization constant, i.e., $\int\hat{p}_{\bY}d\by=C$. Let ${q}_{\bY,\btheta}$ be a PDF model given by KRnet. To train the KRnet, we consider the KL divergence between ${q}_{\bY,\btheta}$ and $p_{\bY}$, where the unknown constant $C$ will shown up as a shift that does not affect the minimization, i.e.,
\begin{equation}\label{eqn:KL_shifted}
D_{\mathsf{KL}}(q_{\bY,\btheta}\|p_{\bY})=\int q_{\bY,\btheta}\log\frac{q_{\bY,\btheta}}{\hat{p}_{\bY}}d\by + \log C=\mathcal{D}_{\btheta}^{\mathsf{pdf}}(\hat{p}_{\bY})+\log C.
\end{equation} 
Minimizing $D_{\mathsf{KL}}(q_{\bY,\btheta}\|p_{\bY})$ is equivalent to minimizing $\mathcal{D}_{\btheta}^{\mathsf{pdf}}$. In general, $\mathcal{D}_{\btheta}^{\mathsf{pdf}}$ needs to be approximated by sampling, which is trivial thanks to the generative modeling. Noting that KRnet corresponds to an invertible mapping $\bZ=f(\bY)$ such that $\bZ\sim\mathcal{N}(0,\mathbf{I})$, we can easily apply the reparameterization trick, i.e.,
\begin{equation}\label{eqn:approx_shifted_KL_KRnet}
\mathcal{D}_{\btheta}^{\mathsf{pdf}}(\hat{p}_{\bY})=\int q_{\bZ}\log\frac{q_{\bY,\btheta}(f^{-1}(\bz))}{\hat{p}_{\bY}(f^{-1}(\bz))}d\bz\approx \frac{1}{N}\sum_{i=1}^N\log\frac{q_{\bY,\btheta}(f^{-1}(\bz^{(i)}))}{\hat{p}_{\bY}(f^{-1}(\bz^{(i)}))}.
\end{equation}
So the training set is simply a set $\{\bz^{(i)}\}_{i=1}^N$ of samples from $\mathcal{N}(\boldsymbol{0},\mathbf{I})$. 

\subsubsection{VAE-KRnet}\label{sec:var_bayes}
If dimension reduction is considered, the latent random variable $\bX$ is taken into account such that the PDF model is given by VAE-KRnet. For this case, let us specify $p_{\bY}$ and $p_{\bX|\bY}$ and use $q_{\bY|\bX}$ to indicate the approximation of $p_{\bY|\bX}$. We then minimize the following objective function 
\begin{equation}\label{eqn:PDF_vae_krnet_mutual}
-D_{\mathsf{KL}}(q_{\bY|\bX,\btheta}p_{\bX,\btheta}\|p_{\bX,\btheta}p_{\bY})+\lambda D_{\mathsf{KL}}(q_{\bY|\bX,\btheta}p_{\bX,\btheta}\|p_{\bX|\bY,\btheta}p_{\bY}),
\end{equation}
where the first term corresponds to the mutual information between $\bX$ and $\bY$, and 
$\lambda$ is the Lagrange multiplier of the constraint $D_{\mathsf{KL}}(q_{\bY|\bX,\btheta}p_{\bX,\btheta}\|p_{\bX|\bY,\btheta}p_{\bY})=0$. The second term here acts as a regularization term with $\lambda>0$. Minimizing the given objective function will maximize the mutual information between $\bX$ and $\bY$ subject to the constraint that $D_{\mathsf{KL}}(q_{\bY|\bX,\btheta}p_{\bX,\btheta}\|p_{\bX|\bY,\btheta}p_{\bY})$ is small as much as possible. Removing the normalization constant in $p_{\bY}$ as in the previous section, we define
\begin{align}
&\mathcal{E}_{\btheta}^{\mathsf{pdf}}(\hat{p}_{\bY})\nonumber\\
=&\int q_{\bY|\bX,\btheta}p_{\bX,\btheta}\left[\log\left(\frac{q_{\bY|\bX,\btheta}p_{\bX,\btheta}}{\hat{p}_{\bY}}\right)^{\lambda-1}+\log p_{\bX,\btheta}-\log p^{\lambda}_{\bX|\bY,\btheta}\right]d\bx d\by,\label{eqn:KL_knwon_PDF}
\end{align}
where $\lambda>1$. When we decrease $\lambda$ from $\infty$, the term $\log\left(\frac{q_{\bY|\bX,\btheta}p_{\bX,\btheta}}{\hat{p}_{\bY}}\right)^{\lambda-1}=0$ at $\lambda=1$. Then $\hat{p}_{\bY}$ disappears from the loss function and the problem becomes ill-posed, meaning that the minimum will be $-\infty$. When $\lambda<1$, the regularization term is even weaker, and the problem will be still ill-posed. Similar to equation \eqref{eqn:approx_shifted_KL_KRnet}, we can approximate $\mathcal{E}_{\btheta}^{\mathsf{pdf}}(\hat{p}_{\bY})$ using the reparameterization trick.

When $\lambda$ goes to infinity, minimizing the objective equation \eqref{eqn:PDF_vae_krnet_mutual} is equivalent to minimize directly the KL divergence $D_{\mathsf{KL}}(q_{\bY|\bX,\btheta}p_{\bX,\btheta}\|p_{\bX|\bY,\btheta}p_{\bY})$. It is seen $p_{\bY}=\mathbb{E}_{p_{\bX,\btheta}}[q_{\bY|\bX,\btheta}]$ as long as $D_{\mathsf{KL}}(q_{\bY|\bX,\btheta}p_{\bX,\btheta}\|p_{\bX|\bY,\btheta}p_{\bY})=0$. The main drawback is that this strategy may underestimate the variance although it may predict the mean very well. This is a common problem for variational Bayes especially when the density model is not sufficiently accurate.  {We include a mutual information term in equation \eqref{eqn:PDF_vae_krnet_mutual} to alleviate this issue. Since the joint PDF of $\bY$ and $\bX$ will be defined as $q_{\bY|\bX,\btheta}p_{\bX,\btheta}$ for sample generation, we adjust the two models $q_{\bY|\bX,\btheta}$ and $p_{\bX,\btheta}$ to minimize the uncertainty of $\bY$ after $\bX$ is obtained, i.e., maximizing the mutual information between $\bY$ and $\bX$ using the joint PDF induced by $q_{\bY|\bX,\btheta}p_{\bX,\btheta}$. This is possible because KRnet provides a large family of prior distributions. Numerical experiments show that maximizing the mutual information between $\bX$ and $\bY$ is able to improve the estimation of the variance for properly chosen $\lambda$, which implies that more information from the region of low density can be kept by increasing the weight of the mutual information term.} In general, we obtain the best prediction of the mean at $\lambda=\infty$, and the best prediction of the variance at a finite $\lambda$. This will be demonstrated later by numerical experiments.

\section{Numerical experiments}\label{sec:numerical}
In this section, we examine VAE-KRnet by some numerical experiments. All algorithms are implemented by Tensorflow 2 and the optimization solver is chosen as ADAM with a  learning rate 1e-3 \cite{ADAM_2017}. All neural networks used in equation \eqref{eqn:NN}, encoder and decoder have fully connected hidden layers. For simplicity, the neural networks for both encoder and decoder have the same configuration. The neural networks for $f_{\mathsf{pr}}(\cdot)$ and $f_{\mathsf{en}}(\cdot)$ differs only with respect to the depth or the number of the general coupling layers $f_{k,i}^{\mathsf{inner}}$. In KRnet, the dimension will be reduced one by one if a specification is not given explicitly. No nonlinear invertible layers and rotation layers will be used. We specify some parameters: $D$: the number of hidden layers for both encoder and decoder, $N_D$: the number of neurons for each hidden layer in the encoder and decoder, $L_{\mathsf{pr}}$: the number of general coupling layers in  $f_{\mathsf{pr}}(\cdot)$ (see figure \ref{fig:structure_diagram}), $L_{\mathsf{en}}$: the number of general coupling layers in  $f_{\mathsf{en}}(\cdot)$, and $N_L$: the number of neurons for the neural network in equation \eqref{eqn:NN}. The training set has $10^5$ samples. Four minibatches are used for the estimation of data distribution, and the whole training set is used for the estimation of the posterior. The validation set has 2e5 samples whenever needed. The validation set is large such that the integration errors for the computation of statistics can be ignored compared to the errors of the model. 

\subsection{The linear model}
We first consider the linear model \eqref{eqn:linear_reduction}, where we assume that the column vectors of $\mbA$ are sampled from $\mathcal{N}(0,\mathbf{I})$ subject to $\ell_2$ normalization. When $\sigma$ is small, the distribution of $\bY$ is mainly a $d$-dimensional distribution of $\bX\in\mathbb{R}^d$, which is embedded in a $n$-dimensional space. For computation, $\bY\in\mathbb{R}^{10}$, and $\sigma=0.1$. 

For a prescribed prior $p_{\bX}$, we generate the samples of $\bY$ from the linear model \eqref{eqn:linear_reduction} to form a training set. We will measure the performance of the model using the following quantity (see equation \eqref{eqn:q_to_p})
\begin{equation}
\delta_{\btheta,\bphi,\bbeta,\balpha}=\mathbb{E}_{p_{\bY}}\left[D_{\mathsf{KL}}(q_{\bX|\bY,\bphi,\balpha}\|p_{\bX|\bY,\btheta,\bbeta})\right]=-\mathbb{E}_{p_{\bY}}\left[\mathcal{L}^{\bX}_{\btheta,\bphi,\bbeta,\balpha}(\bY)\right]-h(\bY),
\end{equation}
where $h(\bY)$ is the differential entropy of $p_{\bY}$. Assuming that $p_{\bX,\bbeta}$ and $p_{\bY|\bX,\btheta}$ cover the true prior $p_{\bX,\mathsf{true}}$ and the true likelihood $p_{\bY|\bX,\mathsf{true}}$, $\delta_{\btheta,\bphi,\bbeta,\balpha}=0$ if $q_{\bX|\bY,\bphi,\balpha}$ is able to recover $p_{\bX|\bY,\btheta,\bbeta}$ induced by $p_{\bX,\bbeta}$ and $p_{\bY|\bX,\btheta}$.  Based on the definition of the linear model, we have
\[
h(\bY)=-\mathbb{E}_{p_{\bY}}\left[\log p_{\bY}\right]=-\mathbb{E}_{p_{\bY}}\left[\log \mathbb{E}_{p_{\bX}}[p_{\bY|\bX}]\right], 
\]
which can be computed at the pre-processing stage. $\mathbb{E}_{p_{\bY}}\left[\mathcal{L}^{\bX}_{\btheta,\bphi,\bbeta,\balpha}(\bY)\right]$  
will be approximated as 
\[
\mathbb{E}_{p_{\bY}}\left[\mathcal{L}^{\bX}_{\btheta,\bphi,\bbeta,\balpha}(\bY)\right]\approx\frac{1}{N}\sum_{i=1}^N
\mathcal{L}^{\bX}_{\btheta,\bphi,\bbeta,\balpha}(\by^{(i)}),
\]
where $\{\by^{(i)}\}_{i=1}^N$ is a validation set which is independent of the training set. 

\subsubsection{A Gaussian prior}
\begin{figure}	\center{
		\includegraphics[width=0.6\textwidth]{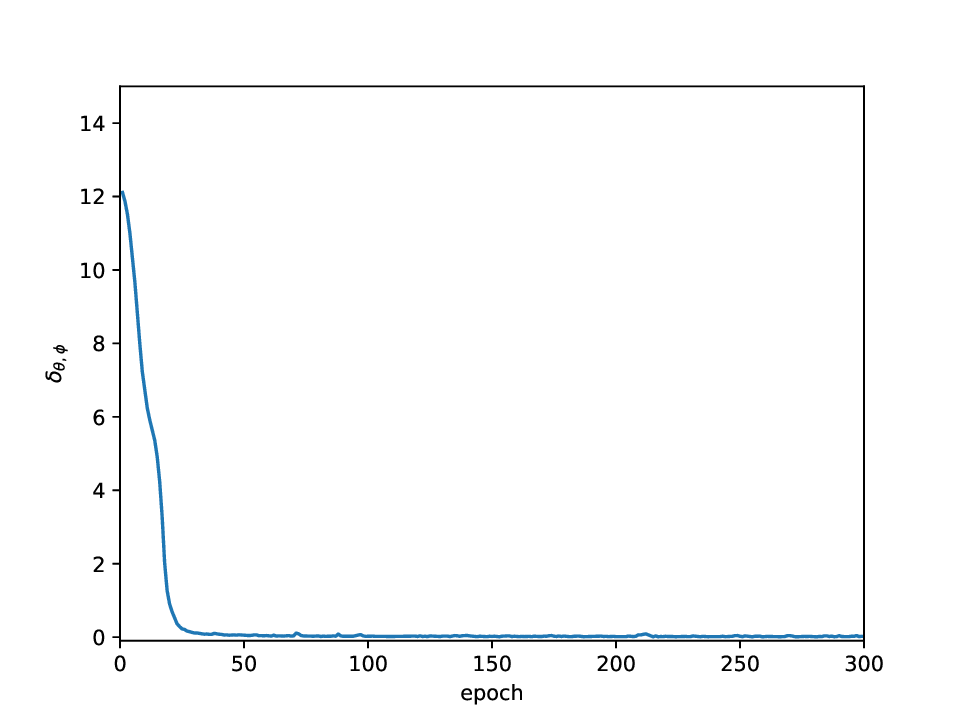}
	}
	\caption{The convergence behavior of VAE for the linear model with a Gaussian prior.}\label{fig:Gaussian_prior_VAE}
\end{figure}

Let $\bX\in\mathbb{R}^2$, 
\[
p_{\bX}(\bx)=\mathcal{N}(0,\mathbf{\Sigma}_{\bX}),\quad p_{\bY|\bX}(\by|\bx)=\mathcal{N}(\mbA\bx,\sigma^2\mathbf{I}).
\]
We have
\[
p_{\bY}=\mathcal{N}(0,\sigma^2\mathbf{I}+\mbA\mathbf{\Sigma}_{\bX}\mbA^{\mathsf{T}}),
\]
which yields that
\[
h(\bY)=5(1+\log(2\pi))+\frac{1}{2}\log\left|\sigma^2\mathbf{I}+\mbA\mathbf{\Sigma}_{\bX}\mbA^{\mathsf{T}}\right|.
\]
Let $\mathbf{\Sigma}_{\bX}=\mathbf{I}$. We sample the two column vectors of $\mbA$ from $\mathcal{N}(0,\mathbf{I})$ then normalize them. We know from section \ref{sec:VAE_linear} that the posterior can be recovered by the canonical VAE subject to a rotation of $\bX$. We let $D=2$ and $N_D=32$. We add one scaling and bias layer after each hidden layer, see section \ref{sec:layers_krnet}, to improve the efficiency. The convergence behavior is shown in figure \ref{fig:Gaussian_prior_VAE}, where a fast decay to zero has been observed, indicating that the model has been exactly recovered. 

\subsubsection{A 2d Gaussian prior with a hole}
We now look at a 2d non-Gaussian prior. We assume that $\bX\sim \mathcal{N}(0,\mathbf{I})$. To introduce correlation between $X_1$ and $X_2$, we consider the data satisfying 
\[
B=\left\{\bx|\|\mathbf{R}^{\alpha,\theta}\bx\|_2\geq C\right\},
\]
where $0<C<\infty$, and $\mathbf{R}$ is a matrix defined as
\[
\mathbf{R}^{\alpha,\theta}=\left[
\begin{array}{cc}
\alpha & 0\\
0&1
\end{array}
\right]\left[
\begin{array}{cc}
\cos\theta& -\sin\theta\\
\sin\theta&\cos\theta
\end{array}
\right],
\]
corresponding to a rotation and a stretch. Simply speaking, the distribution is given by 2d standard Gaussian subject to an elliptic hole.  We let $\alpha=3.0$ and $\theta=\pi/4$. To this end, we have prescribed  
\[
p_{\bX}(\bx)=\frac{I_B(\bx)p_{X_1}(x_1)p_{X_2}(x_2)}{\int_{\mathbb{R}^2}I_B(\bx)p_{X_1}(x_1)p_{X_2}(x_2)dx_1dx_2},\quad p_{\bY|\bX}(\by|\bx)=\mathcal{N}(\mbA\bx,\sigma^2\mathbf{I}),
\]
where $p_{X_i}=\mathcal{N}(0,1)$ and $I_B(\bx)$ is an indicator function.

\begin{figure}	\center{
		\includegraphics[width=0.6\textwidth]{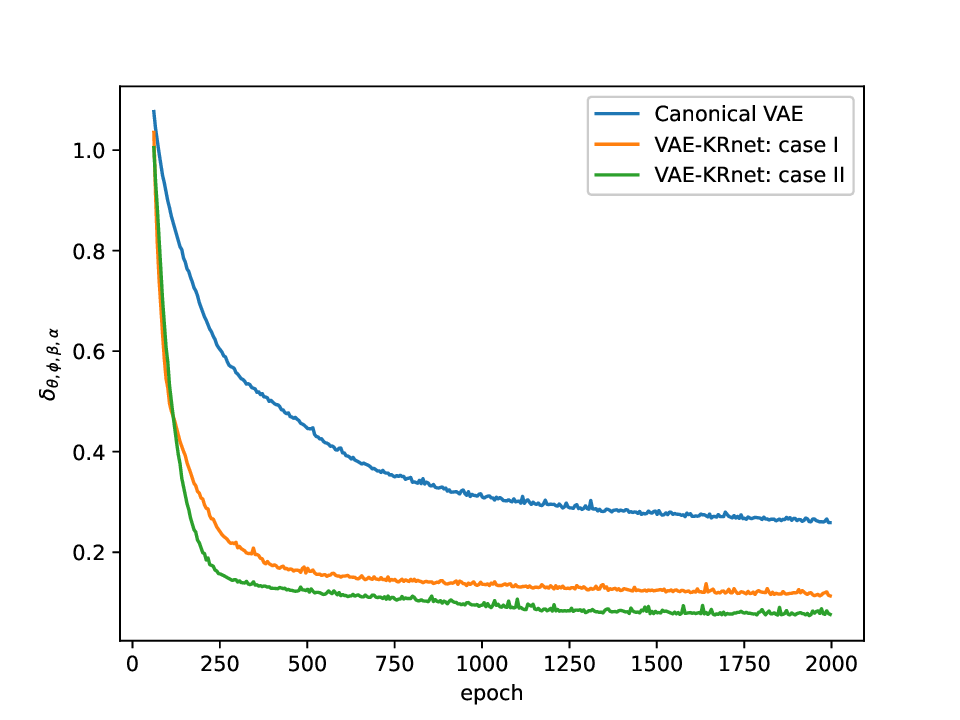}
	}
	\caption{The convergence behavior of VAE and VAE-KRnet for the linear model with a non-Gaussian prior. VAE-KRnet I has a generalized prior, and VAE-KRnet II has both generalized prior and posterior. }\label{fig:Gaussian_w_hole_prior_VAE_KRnet}
\end{figure}

We first show the effect of the generalized prior and posterior. We consider three models: canonical VAE, VAE-KRnet I, and VAE-KRnet II, where VAE-Krnet I has a generalized prior and VAE-KRnet II has both generalized prior and posterior. For the sake of comparison we consider simple configurations. We let $D=1$, $L_{\mathsf{pr}}=L_{\mathsf{en}}=2$, $N_D=32$, and $N_L=24$, whenever the corresponding components are needed in the model. The convergence behavior of these three models has been shown in figure \ref{fig:Gaussian_w_hole_prior_VAE_KRnet}. It is seen that both the generalized prior and posterior are able to improve the performance of the canonical VAE. It appears that $f_{\mathsf{pr}}(\cdot)$ can improve the performance more effectively than $f_{\mathsf{en}}(\cdot)$. 

\begin{figure}	\center{
		\includegraphics[width=0.9\textwidth]{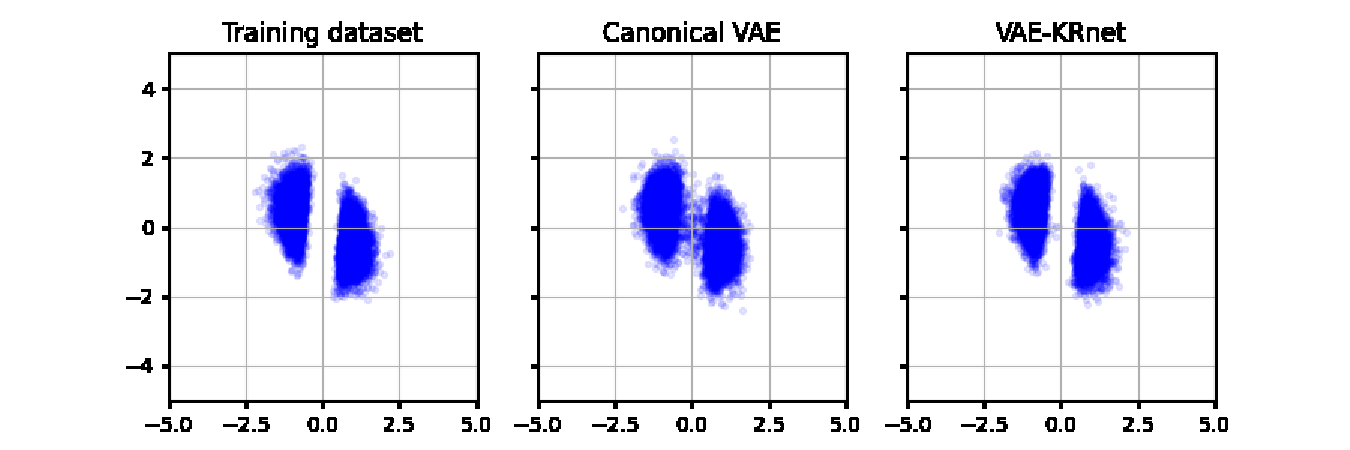}
	}
	\caption{The distribution of $(Y_6,Y_9)$. Left: training set; Middle: samples generated by the canonical VAE; Right: samples generated by VAE-KRnet.}\label{fig:compare_distirubtion_linear_model}
\end{figure}

\begin{figure}	\center{
		\includegraphics[width=0.6\textwidth]{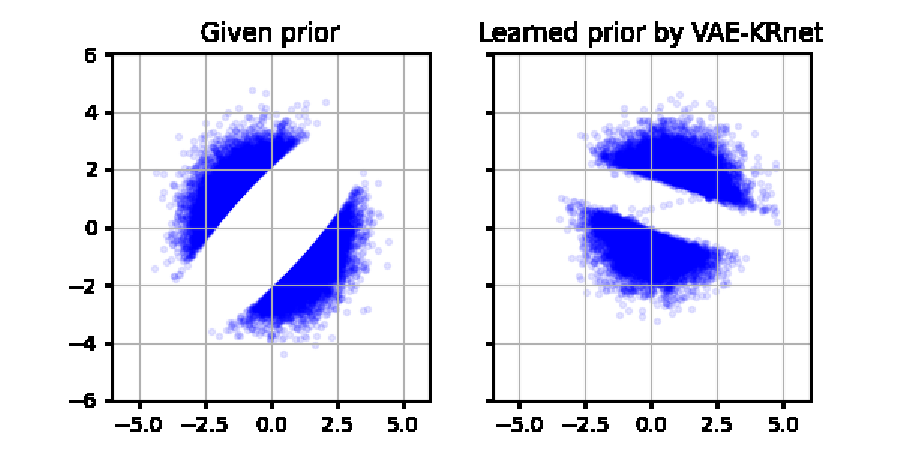}
	}
	\caption{Left: the given prior distribution; Right: the learned prior distribution by VAE-KRnet.}\label{fig:Gaussian_w_hole_simulated_prior_VAE_KRnet}
\end{figure}

We now compare the simulated distributions of $\bY$ given by canonical VAE and VAE-KRnet II, where we let $D=2$, $L_{\mathsf{pr}}=8$, $L_{\mathsf{en}}=2$, $N_D=32$ and $N_L=24$.  The results have been shown in figure \ref{fig:compare_distirubtion_linear_model}. It is seen that canonical VAE is effective to capture the main structure of the distribution while VAE-KRnet is able to capture more details than canonical VAE. 
In figure \ref{fig:Gaussian_w_hole_simulated_prior_VAE_KRnet}, we compare the  given prior and the learned prior by VAE-KRnet. It is seen that the learned prior shares some similarities with the given prior. Note that any invertible mapping of $\bX$ provides a latent variable. Therefore we do not expect the learned prior is the same as the prescribed one. 

\subsubsection{A 3d Gaussian prior with holes}
\begin{figure}	\center{
		\includegraphics[width=0.6\textwidth]{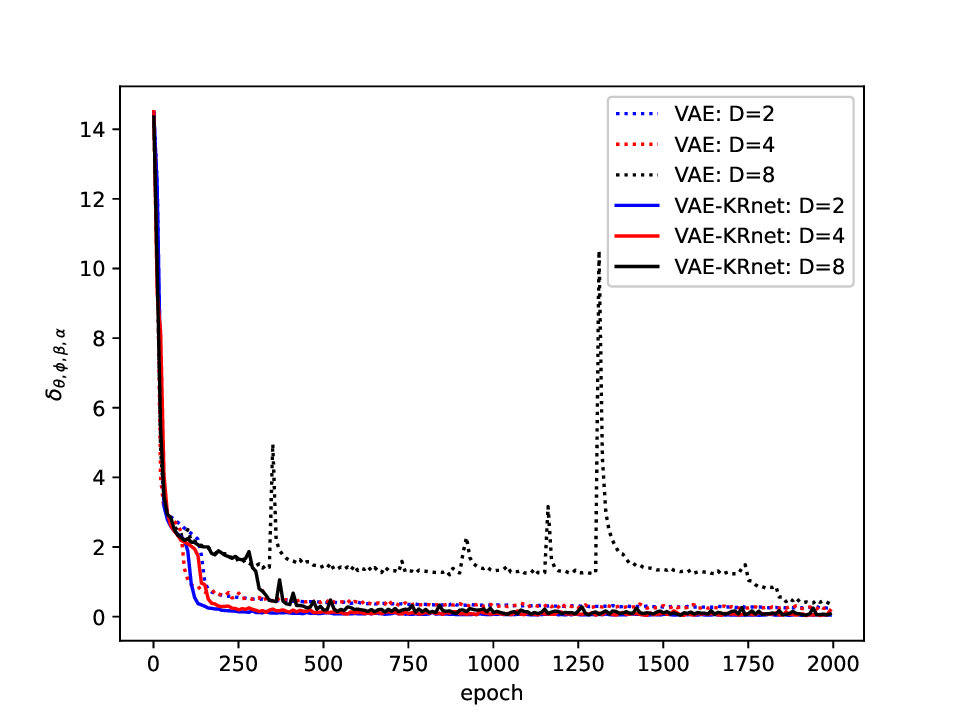}
	}
	\caption{The convergence behavior of VAE and VAE-KRnet in terms of $D$.}\label{fig:Gaussian_w_hole_prior_cong_3d}
\end{figure}

\begin{figure}	\center{
		\includegraphics[width=0.6\textwidth]{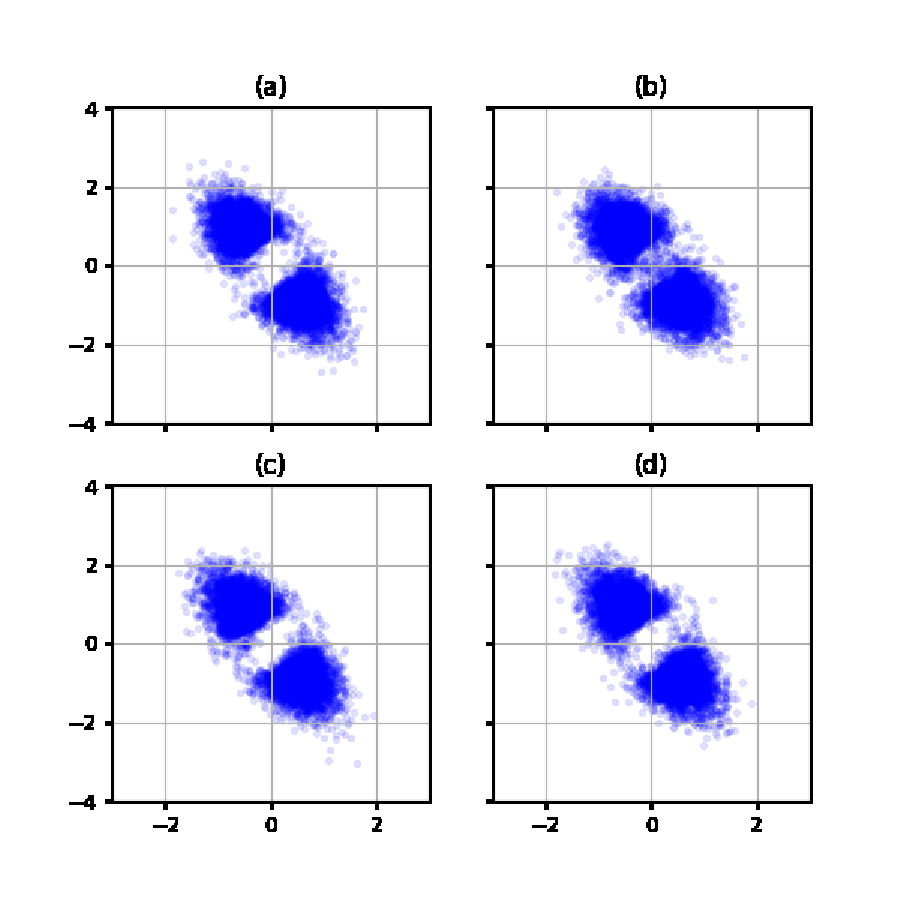}
	}
	\caption{The samples of $(Y_6,Y_8)$  given by VAE and VAE-KRnet with $D=2$. (a): training set; (b): canonical VAE; (c): VAE-KRnet with $L_{\mathsf{pr}}=8$; (d): VAE-KRnet with $L_{\mathsf{pr}}=10$.}\label{fig:Gaussian_w_hole_prior_compare_samples_3d}
\end{figure}

\begin{figure}	\center{
		\includegraphics[width=0.6\textwidth]{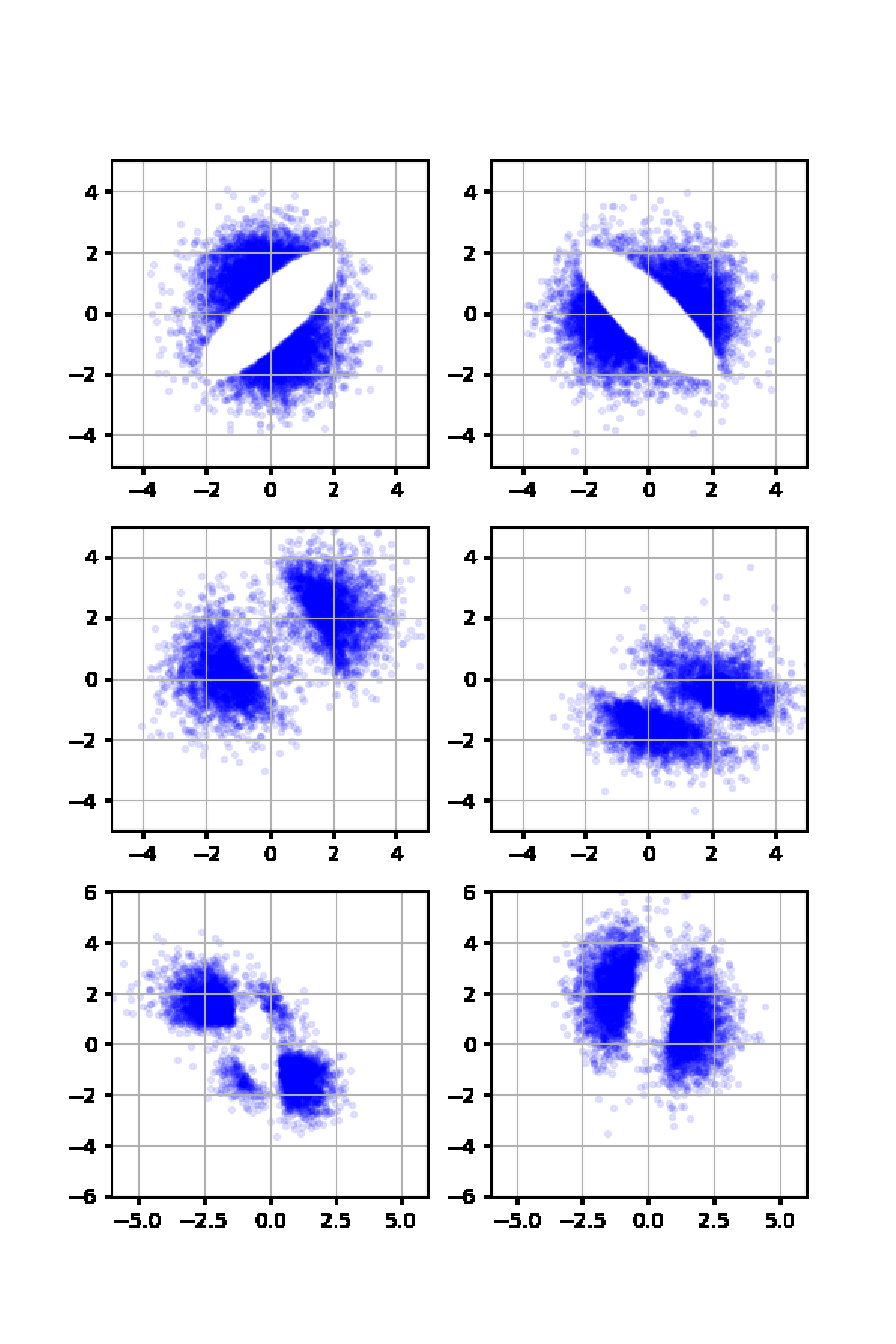}
	}
	\caption{The prescribed and learned prior distributions. For each row, the left plot gives the distribution of $(X_1,X_2)$, and the right plot corresponds to $(X_2,X_3)$. From top to bottom, the first row  corresponds to the prescribed prior distribution, and the second and third rows correspond to the learned prior distributions given by VAE-KRnet with the same configuration except that  $L_{\mathsf{pr}}=8,10$ respectively. }\label{fig:Gaussian_w_hole_simulated_prior_VAE_KRnet_3d}
\end{figure}

We now consider a case that $\bX\in\mathbb{R}^3$. For $\bx=[x_1,x_2,x_3]^\mathsf{T}$, we let $\bx_i=[x_i,x_{i+1}]^\mathsf{T}$, $i=1,2$, which includes two adjacent components of $\bx$. We sample from $\bX\sim\mathcal{N}(0,\mathbf{I})$ and keep the data 
\[
B=\left\{\bx_i|\|\mathbf{R}^{\alpha,\theta_i}\bx_i\|_2\geq C\right\},\quad i=1,2.
\]
In other words, for any two adjacent dimensions we generate an elliptic hole. We let $\alpha=3$, $\theta_1=\pi/4$ and $\theta_2=3\pi/4$ (see the top two plots in figure \ref{fig:Gaussian_w_hole_simulated_prior_VAE_KRnet_3d}). We first check the performance of canonical VAE and VAE-KRnet with respect to $D$, the depth of the neural networks for the encoder and decoder. We let $L_{\mathsf{pr}}=8$, $L_{\mathsf{en}}=2$, $N_L=24$ and $N_D=32$. The convergence behavior has been plotted in figure \ref{fig:Gaussian_w_hole_prior_cong_3d}. 
First of all, VAE-KRnet has a better performance. For a fixed $D$, VAE-KRnet reaches a smaller loss than VAE. Second, VAE-KRnet is more robust than VAE. When $D=8$, VAE has been stuck in a local minimizer until epoch $\approx$1750 before it goes to a better local minimizer. The introduction of generalized prior and posterior makes it much easier to escape the basin of attraction of such a local minimizer. It is seen that VAE-KRnet with $D=8$ is able to achieve the same loss as other configurations when the epoch is about 500 although the degree of fluctuation is bigger due to the increased model complexity. In figure \ref{fig:Gaussian_w_hole_prior_compare_samples_3d}, we compare the distributions simulated by VAE and VAE-KRnet. It is seen that much more details can be captured by VAE-KRnet than VAE. In figure \ref{fig:Gaussian_w_hole_simulated_prior_VAE_KRnet_3d}, we plotted the prior distributions given by VAE-KRnet, where the only difference in configuration is that $L_{\mathsf{pr}}=8,10$. It is seen that the learned prior distributions are quite different although the two configurations of VAE-KRnet are similar and yield almost the same approximation of the data distribution. 


\subsection{A linear Bayesian inverse problem}

We consider the following linear model for the inverse problem:
\begin{equation}\label{eqn:linear_inv}
\hat{\bY}=\mathbf{K}\bY+\bxi,
\end{equation}
where $\mathbf{K}\in\mathbb{R}^{k\times n}$, and $\bxi\sim\mathcal{N}(0,\sigma^2\mathbf{I})$. In particular, we assume that $\mathbf{K}$ is ill-conditioned in the sense that its singular values decays fast. Assume a Gaussian prior $\mathcal{N}({\bmu}_{\mathsf{pr}},\mathbf{\Sigma}_{\mathsf{pr}})$ is used. The posterior is
\begin{equation}\label{eqn:inv_pdf_unscaled}
p_{\mathsf{post}}(\by)\propto \hat{p}_{\mathsf{post}}(\by)=\exp\left(-\frac{|\hat{\by}-\mathbf{K}\by|^2}{2\sigma^2}\right)\exp\left(-\frac{1}{2}|\by-{\bmu}_{\mathsf{pr}}|^2_{\mathbf{\Sigma}^{-1}_{\mathsf{pr}}}\right),
\end{equation}
what $\hat{p}_{\mathsf{post}}(\by)$ is the unnormalized posterior such that $p_{\mathsf{post}}(\by)\int \hat{p}_{\mathsf{post}}(\by)d\by=\hat{p}_{\mathsf{post}}(\by)$, and ${|\by|^2_{\mathbf{\Sigma}^{-1}_{\mathsf{pr}}}}=\by^\mathsf{T}\mathbf{\Sigma}_{\mathsf{pr}}^{-1}\by$ defines a weighted $\ell_2$ norm induced by the precision matrix. We want to find a low-dimension latent random variable $\bX\in\mathbb{R}^d$ for $\bY$ such that $\int p_{\bX,\bY}d\bx\approx p_{\mathsf{post}}(\by)$, where $p_{\bX,\bY}$ is the joint PDF of $\bX$ and $\bY$, and will be provided by VAE-KRnet. 

We define problem \eqref{eqn:linear_inv} using an integral equation 
\begin{equation}
g(x)=\int_{\mathbb{R}} K(x,y)f(y)dy,
\end{equation}
where $K(x,y)$ is the kernel of a compact operator that is of trace class, positive and self-adjoint. Let $(\lambda_i,e_i(x))$ indicate the eigen-pairs of $K(x,y)$. Assume that
\begin{equation}\label{eqn:f_i}
f(y)\approx\sum_{i=1}^Mf_ie_i(y).
\end{equation}
We consider the equation 
\begin{equation}
g(x_j)\approx\sum_{i=1}^Mf_i\lambda_ie_i(x_j), \quad j = 1,\ldots, N_x,
\end{equation}
where $x_i$ are collocation points. This yields a linear system 
\begin{equation}
\boldsymbol{g}=\mathbf{E}\mathbf{\Lambda}\boldsymbol{f},
\end{equation}
where $e_{ij}=e_{i}(x_j)$, $g_j=g(x_j)$, $i=1,\ldots,M$, $j=1,\ldots,N_x$. We then let $\mathbf{K}=\mathbf{E\Lambda}$ in equation \eqref{eqn:linear_inv} and $\hat{\by}=\boldsymbol{g}$ is the data. We need to infer the coefficients $f_i$ in equation \eqref{eqn:f_i}. 

For simplicity and without loss of generality, we here consider an artificial case, where we let $e_i(x)=\frac{1}{\sqrt{\pi}}\cos(ix)$ with $x\in[0,2\pi]$, and $\lambda_i=i^{-\gamma}$ with $\gamma>0$. The collocation points are sampled from a uniform distribution on $[0,2\pi]$. This way, the column vectors of $\mathbf{E}$ are nearly mutually orthogonal due to the properties of $\cos(ix)$. The condition number of $\mathbf{K}^\mathsf{T}\mathbf{K}$ depends on the value of $\gamma$, where the eigenvalue $\lambda_i^2$ decays faster for a larger $\gamma$. We define a vector $\by_0$ with $y_{0,i}=i^{-2.0}\sin(i)$, $i=1,\ldots,n$, and generate the data $\hat{\by}=\mathbf{K}\by_0+\sigma\bxi_0$, where $\bxi_0$ is a sample from $\mathcal{N}(0,\mathbf{I})$. We then consider a Bayesian inverse problem \eqref{eqn:linear_inv} using $\hat{\by}$ as the given data. The true posterior is 
\begin{equation}\label{eqn:post_true}
p_{\mathsf{post}}(\by)=\mathcal{N}(\sigma^{-2}(\mathbf{\Sigma}_{\mathsf{pr}}^{-1}+\sigma^{-2}\mbK^\mathsf{T}\mbK)^{-1}\mbK^{\mathsf{T}}\hat{\by}, (\mathbf{\Sigma}_{\mathsf{pr}}^{-1}+\sigma^{-2}\mbK^\mathsf{T}\mbK)^{-1}).
\end{equation}
Since we often choose $\mathbf{\Sigma}_{\mathsf{pr}}$ as a diagonal matrix, the covariance matrix of $p_{\mathsf{post}}(\by)$ is nearly diagonal by the definition of $\mathbf{K}=\mathbf{E}\mathbf{\Lambda}$, which implies that the components of $\bY$ are nearly independent. To consider dimension reduction, correlation should be introduced. We define a matrix $b_{ij}=e^{-|i-j|/\alpha}$ with $\alpha>0$ and $i,j=1,\ldots,n$, and redefine $\mathbf{K}=\mathbf{E}\mathbf{B}\mathbf{\Lambda}$. The parameter $\alpha$ acts as a correlation length. Note that the column vectors of $\mathbf{EB}$ are not nearly mutually orthogonal any more. Letting $\by=\mathbf{0}$, the normalization constant for $\hat{p}_{\mathsf{post}}$ can be computed as
\begin{equation}\label{eqn:C}
C=\frac{\hat{p}_{\mathsf{post}}(\mathbf{0})}{p_{\mathsf{post}}(\mathbf{0})}=\frac{\sqrt{(2\pi)^n|\mathbf{\Sigma}_{\mathsf{post}}|}}{\exp\left(\frac{1}{2\sigma^2}|\hat{\by}|^2+\frac{1}{2}|{\bmu}_{\mathsf{pr}}|^2_{\mathbf{\Sigma}_{\mathsf{pr}}^{-1}}-\frac{1}{2}|{\bmu}_{\mathsf{post}}|^2_{\mathbf{\Sigma}_{\mathsf{post}}^{-1}}\right)},
\end{equation}
where $\bmu_{\mathsf{post}}$ and $\mathbf{\Sigma}_{\mathsf{post}}$ are the mean and covariance matrix of the posterior \eqref{eqn:post_true}. 

We consider several PDF models for the approximation of the posterior. 1) The mean-field variational family, where all random variables are assumed to be mutually independent. For our problem, each dimension will be assumed to be Gaussian; 2) KRnet; and 3) VAE-KRnet. The mean-field variational family is widely used in practice due to its simplicity and efficiency. We do not include VAE here since VAE-KRnet is more robust than VAE. For the mean-field model, we simply use the ADAM method for optimization without taking advantage of the mutual independence like the CAVI algorithm \cite{Blei_2018}. A direct generalization of the mean-field variational model is the mixture of Gaussians, which is not included here. Instead, we consider VAE-KRnet as a generalization of the mixture of Gaussians, since the latent variables for VAE-KRnet is much more general than the latent variable of the Gaussian mixture model. 

Another issue is the computation of statistics. The statistics will be computed using the model that yields the minimum loss with respect to the validation set. Actually, for our experiments the training set is large enough, meaning that we do not observe that the error will increase in terms of the validation set after the optimization iteration has stabilized. 

We let $\sigma=0.05$ and $\mathcal{N}({\bmu}_{\mathsf{pr}},\mathbf{\Sigma}_{\mathsf{pr}})=\mathcal{N}(\mathbf{0},\mathbf{\Lambda}_{\mathsf{pr}})$, where $\mathbf{\Lambda}_{\mathsf{pr}}$ is a diagonal matrix with $\lambda_{\mathsf{pr},i}=i^{-2.5}$, $i=1,\ldots, n$.  For matrix $\mathbf{B}$, we let $b_{ij}=e^{-|i-j|/3.0}$, $i,j=1,\ldots,n$. Note that from equation \eqref{eqn:approx_shifted_KL_KRnet} we have
\[
\mathcal{D}_{\btheta}^{\mathsf{pdf}}(\hat{p}_{\bY})\geq -\log C.
\] 
The lower bound can be computed by equation \eqref{eqn:C}. We will consider two cases when $n=10,50$. For KRnet, we let $L=6$, $K=5$, $N_L=24$ for both $n=10$ and $n=50$. The dimensions will be deactivated by two if $n=10$ and by ten if $n=50$. In other words, the model complexity of KRnet is the same for $n=10,50$. For VAE-KRnet, we let $L_{\mathsf{pr}}=6$, $L_{\mathsf{en}}=2$, $N_L=24$, $N_D=32$ for all cases. In $f_{\mathsf{pr}}(\cdot)$ and $f_{\mathsf{en}}(\cdot)$, the dimensions will be deactivated by two.  Let $r(x;\bY)=\sum_{i=1}^nY_ie_i(x)$. After we approximate the posterior of $\bY$, we compute the $\mathbb{E}[r](x)$ and $\mathrm{Var}(r)(x)$ and compare them to the exact  values. 

\begin{figure}	\center{
\includegraphics[width=0.49\textwidth]{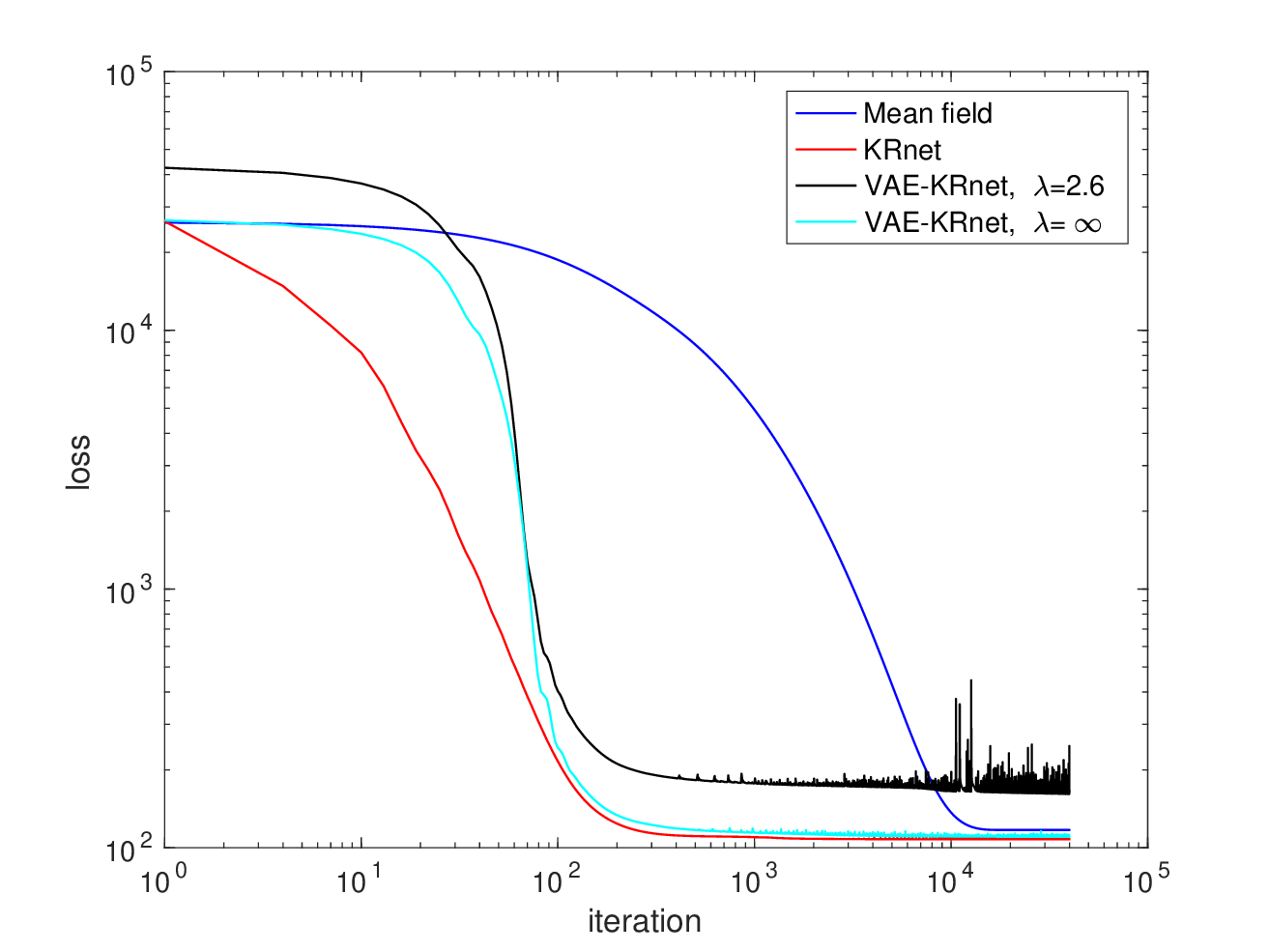}
\includegraphics[width=0.475\textwidth]{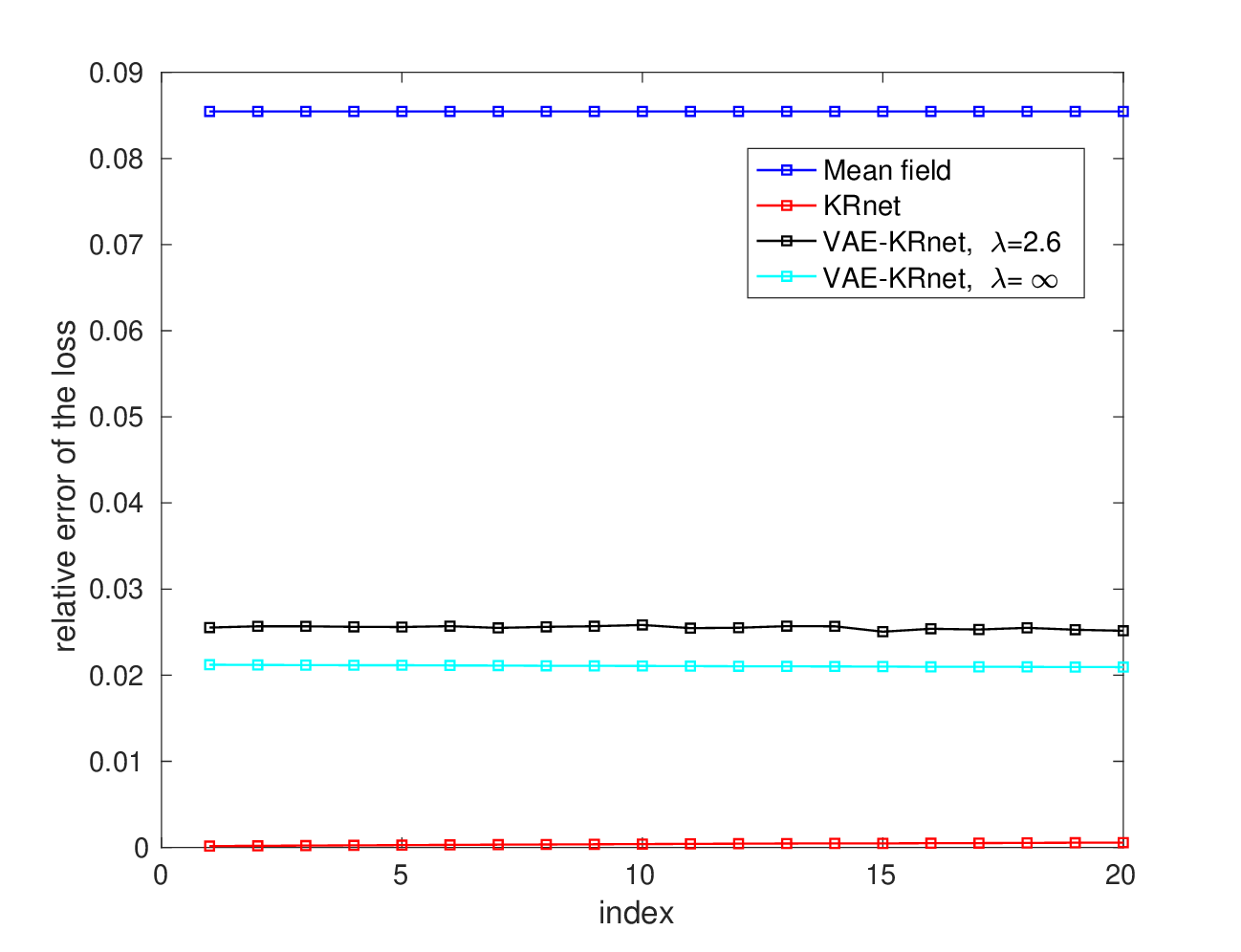}}
	\caption{The evolution behavior of the ADAM method for different PDF models. $n=10$. Left: iterations up to 3e5; Right: iterations from 3e5 to 5e5. Each node corresponds to the minimum loss in every 1000 iterations.}\label{fig:d10_II_cong}
\end{figure}

We first consider a 10-dimensional case, where we have $-\log C=107.94 $ from equation \eqref{eqn:C}. In figure \ref{fig:d10_II_cong} we plotted the evolution behavior of the ADAM method, where the global behavior is given on the left with respect to the loss, and the stabilized behavior is given on the right with respect to the relative error of the loss:
\[
\frac{|\mathcal{D}_{\btheta}^{\mathsf{pdf}}(\hat{p}_{\bY})+\log C|}{|\log C|}.
\]
 Interestingly, all VAE-KRnet models decay much faster than the mean-field variational model although they are much more complicated. Due to the correlation introduced by $\mathbf{B}$, the mean-field variational model becomes stabilized at a larger relative error than other PDF models, where KRnet performs the best and VAE-KRnet yields a smaller loss for a larger $\lambda$. After the iteration number reaches 3e5, we compute the minimum loss within every 1000 iteration with respect to the validation set, and the results are given in the right plot of figure \ref{fig:d10_II_cong}. It is seem that the minimum loss is quite steady although a lot of fluctuations exist in the optimization iteration. 

\begin{figure}	
\center{
\includegraphics[width=0.49\textwidth]{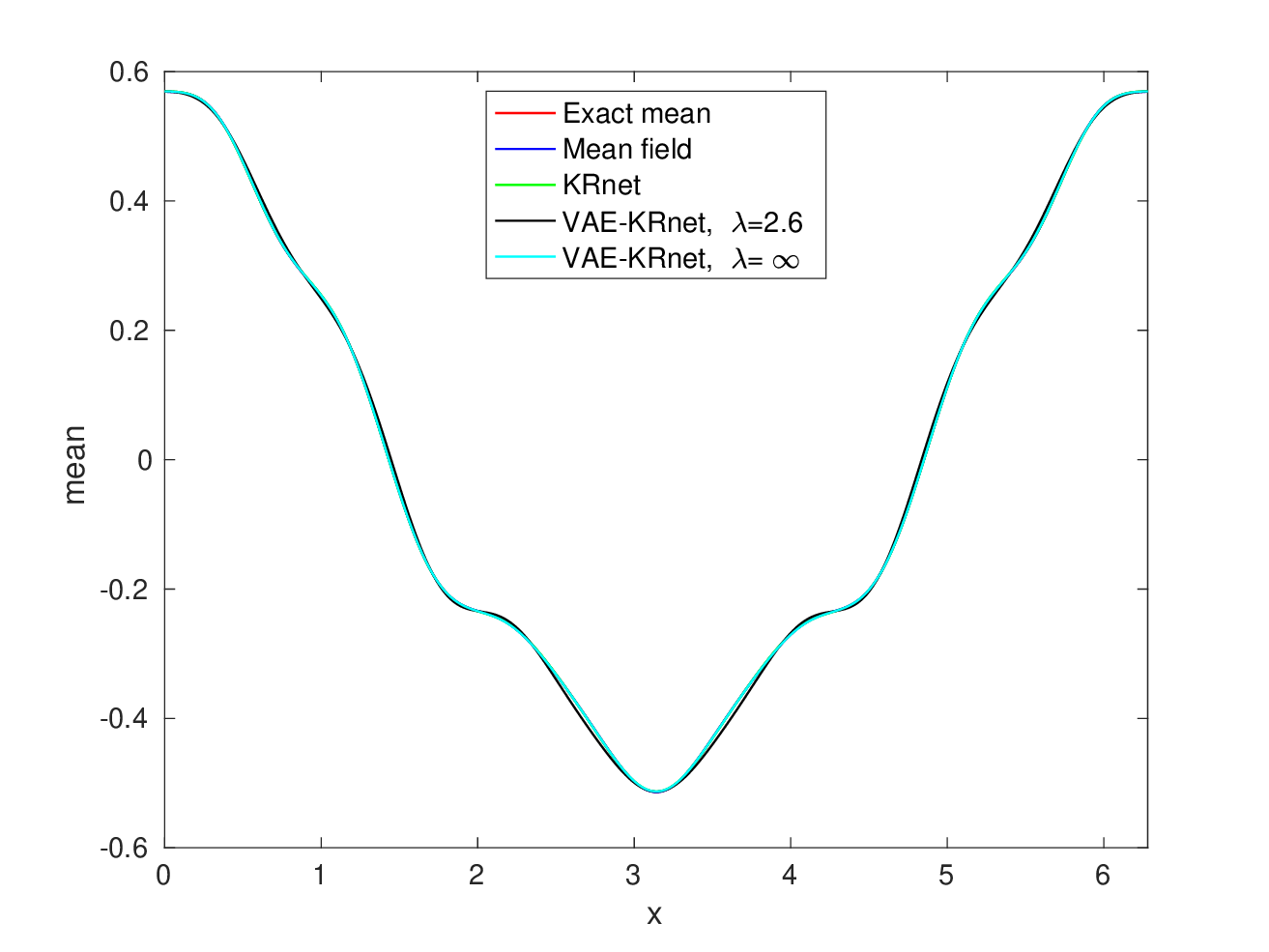}
\includegraphics[width=0.49\textwidth]{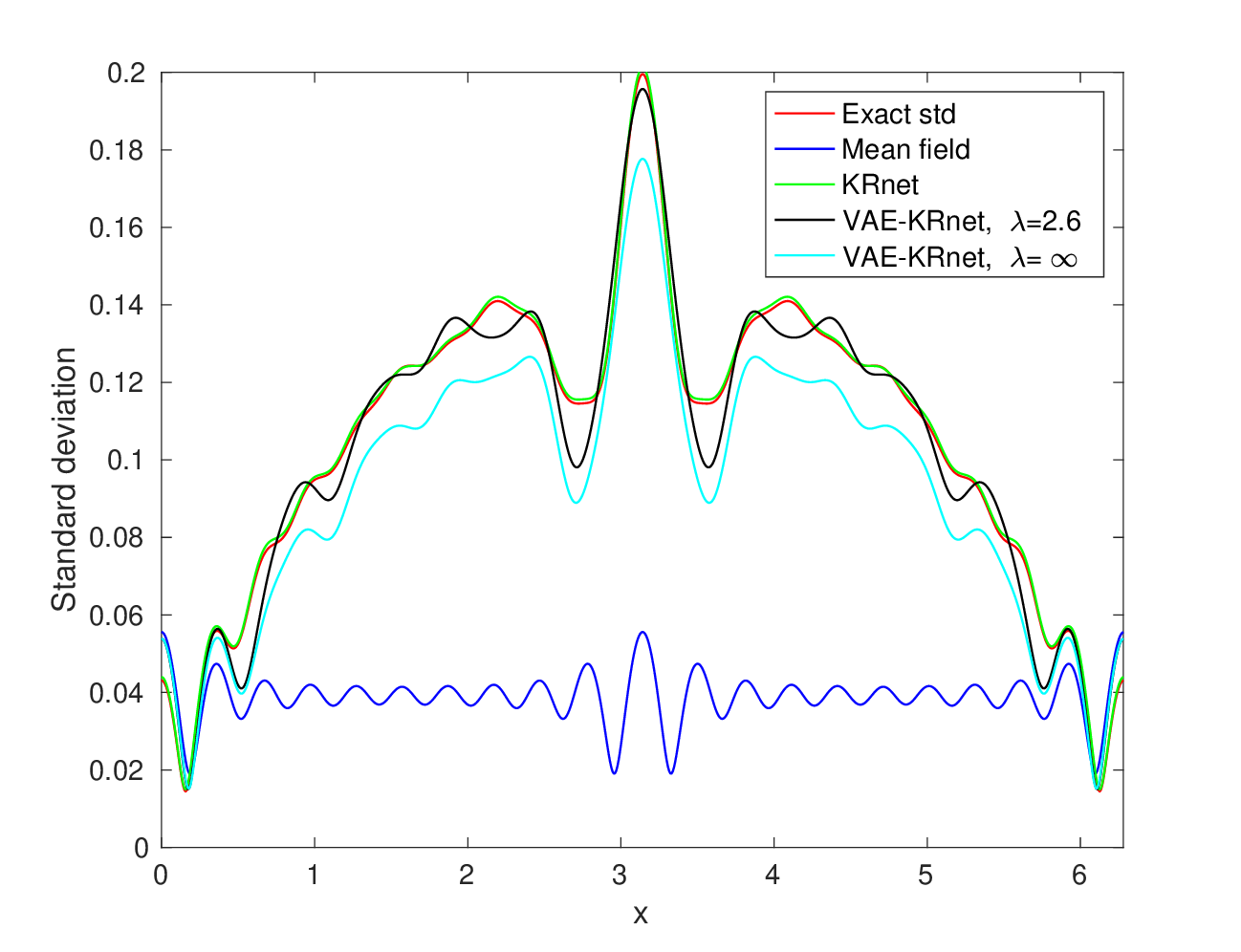}}
	\caption{The statistics given by different PDF models. $n=10$. $d=4$ for VAE-KRnet. Left: Mean; Right: Variance.}\label{fig:d10_II_mean_var}
\end{figure}

\begin{figure}	
\center{\includegraphics[width=0.49\textwidth]{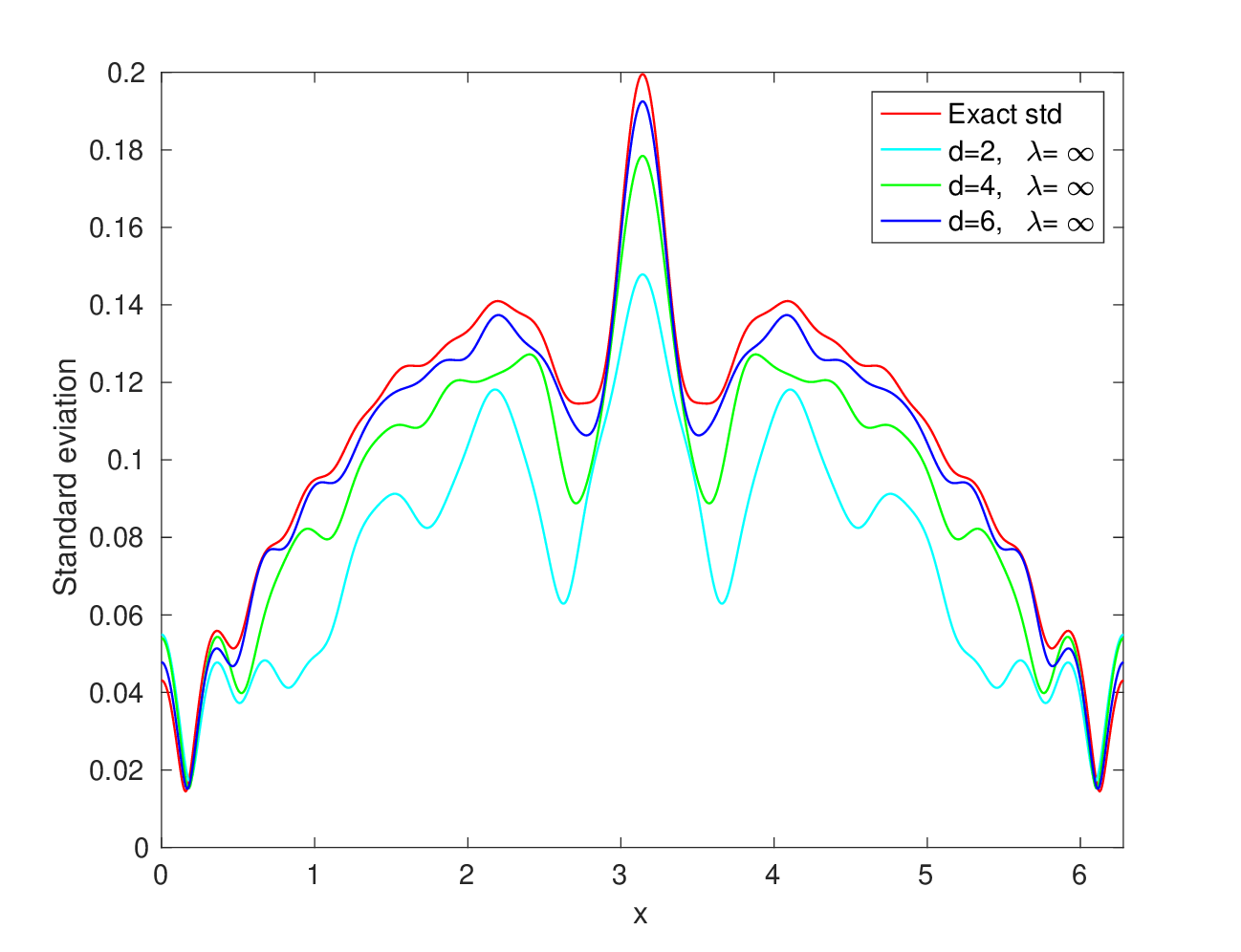}
\includegraphics[width=0.49\textwidth]{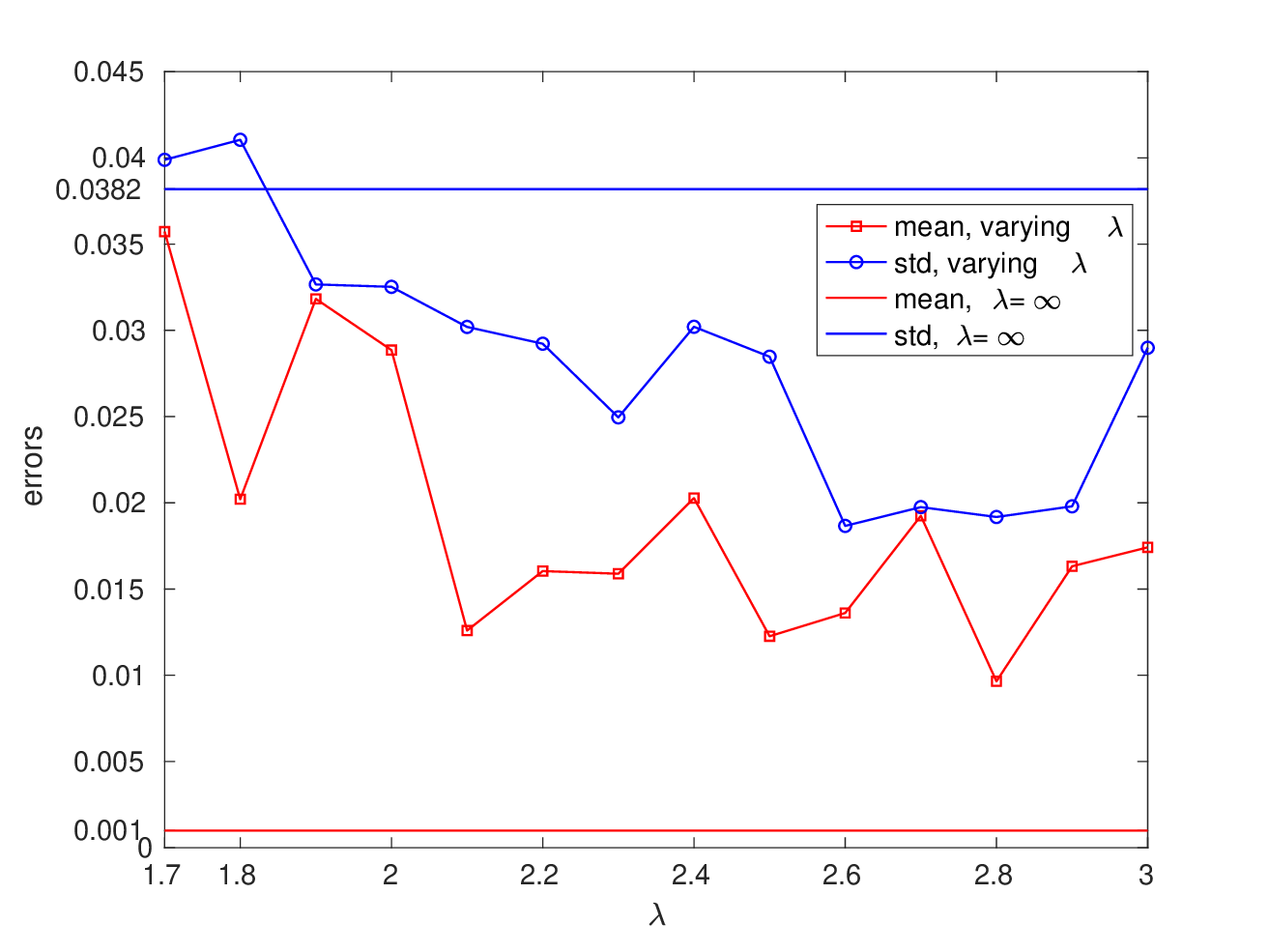}
}
	\caption{VAE-KRnet for different $d$ and $\lambda$. Left: varying $d$ with $\lambda=\infty$. Right: varying $\lambda$ with $d=4$.}
	\label{fig:d10_II_mean_var_mx}
\end{figure}

In figure \ref{fig:d10_II_mean_var}, we plotted the predicted mean and variance. It is seen that the mean is well predicted by all models while the prediction of the variance varies significantly. KRnet yields the best approximation. The mean-field variational model barely captures any characteristics of the variance. VAE-KRnet with $\lambda=2$ yields a better estimation of the variance than $\lambda=\infty$. In figure \ref{fig:d10_II_mean_var_mx} we plotted the effect of the dimension of the latent variable on the left and the effect of the value of $\lambda$ on the right. It is seen that the prediction has been improved by increasing $d$, which is expected. To show the effect of $\lambda$, we compute the errors as follows. Let $\hat{r}(x;\hat{\bY})$ be an approximation of $r(x;\bY)$. We check the following errors:
\[
\frac{\|\mathbb{E}[\hat{r}]-\mathbb{E}[r]\|_{L_2}}{\|\mathbb{E}[r](x)\|_{L_2}},\quad
\frac{\|\mathrm{Var}^{1/2}(\hat{r})-\mathrm{Var}^{1/2}(r)\|_{L_2}}{\|\mathbb{E}[r](x)\|_{L_2}},
\]
for the mean and the standard variation respectively, where the $\|\cdot\|_{L_2}$ is with respect to the spacial variable $x$. {We are interested in a range of $\lambda$, in which the mutual information term in equation \eqref{eqn:PDF_vae_krnet_mutual} helps improve the prediction of the standard deviation. As discussed in section \ref{sec:var_bayes}, when $\lambda\leq 1$ the optimization problem \eqref{eqn:PDF_vae_krnet_mutual} becomes ill-posed. For the problem studied, a smaller error of the standard deviation is observed when we increase $\lambda$ to about 2. As $\lambda$ continues to increase, the error of the standard deviation has an overall trend to decrease until $\lambda$ is about 3, after which we expect that the effect of the mutual information term will become more and more weaker and the error of the standard deviation will increase and approach the error given by $\lambda=\infty$, i.e., the blue horizontal line.} It is seen that VAE-KRnet with $\lambda=\infty$ yields a much smaller error for the mean (the red line) than for the standard deviation (the blue line). Within quite a wide range of $\lambda$, VAE-KRnet with a finite $\lambda$ yields a much more accurate estimation of the variance than VAE-KRnet with $\lambda=\infty$. However, VAE-KRnet with a finite $\lambda$ yields a worse estimation of the mean than VAE-KRnet with $\lambda=\infty$.
 
\begin{figure}	
\center{
\includegraphics[width=0.6\textwidth]{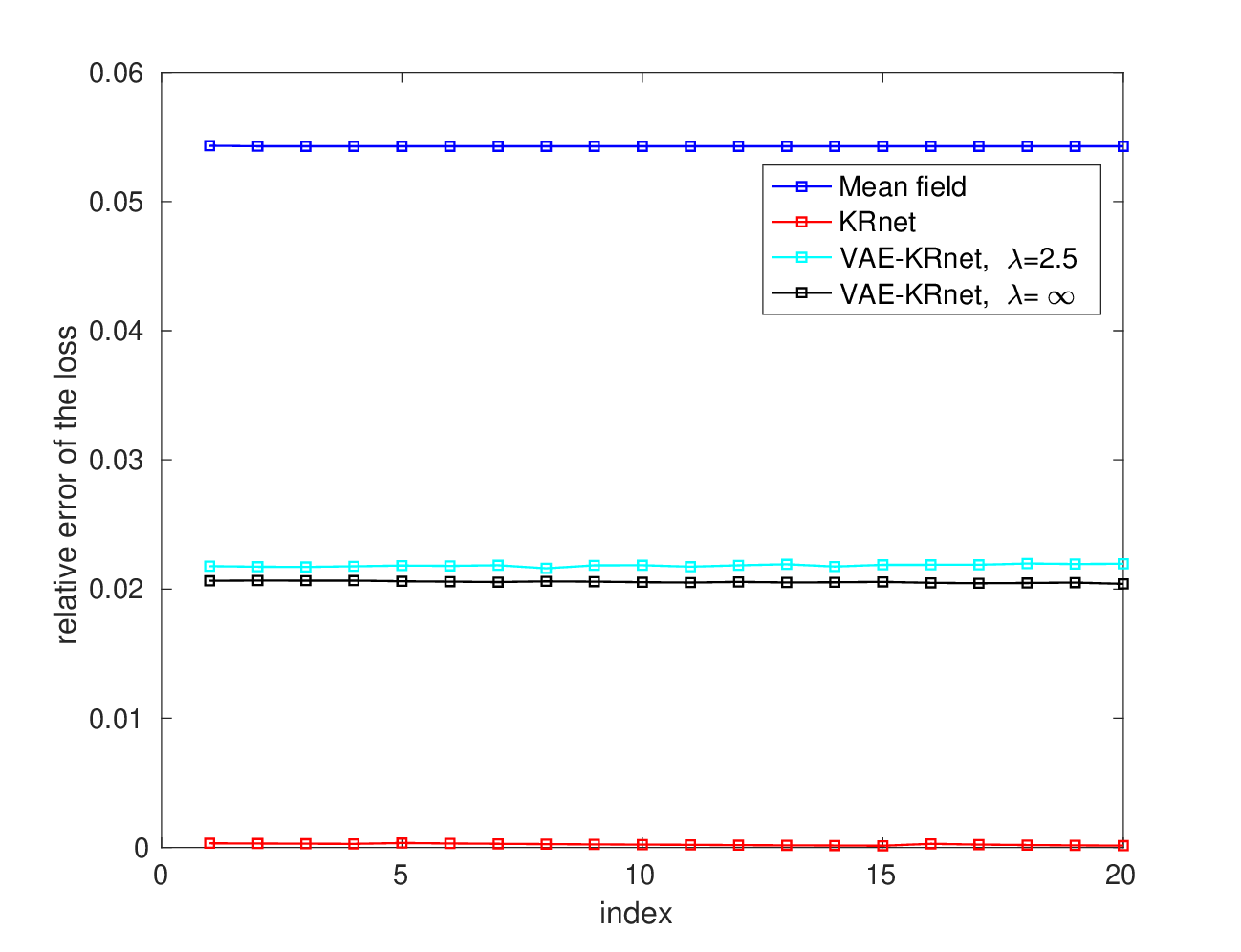}}
	\caption{The stabilized iterations of the ADAM method for different PDF models. $n=50$. $d=8$ for VAE-KRnet. Iterations from 3e5 to 5e5. Each node corresponds to the minimum loss in every 1000 iterations.}\label{fig:d50_II_cong}
\end{figure}

\begin{figure}	
\center{
\includegraphics[width=0.49\textwidth]{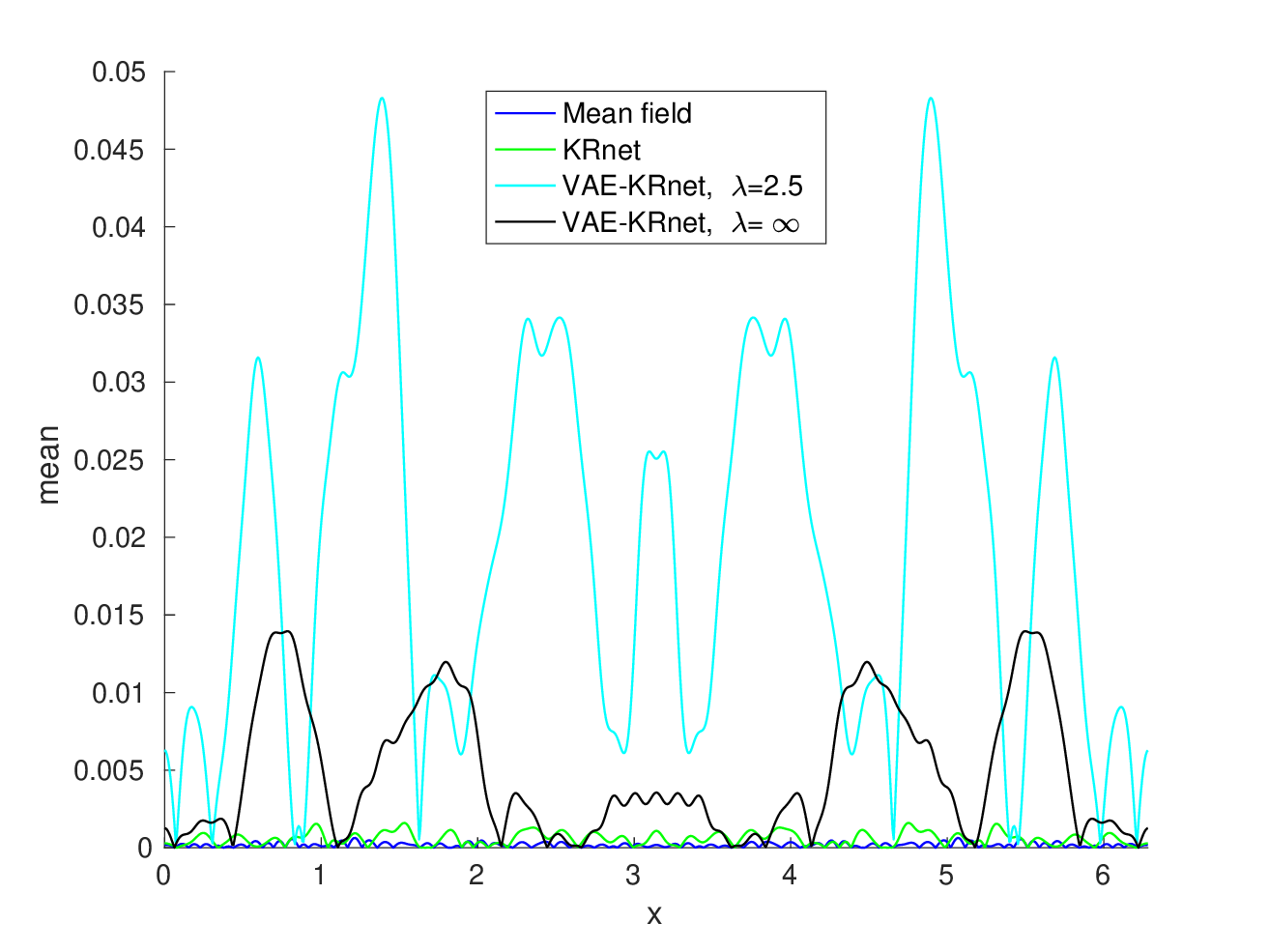}
\includegraphics[width=0.49\textwidth]{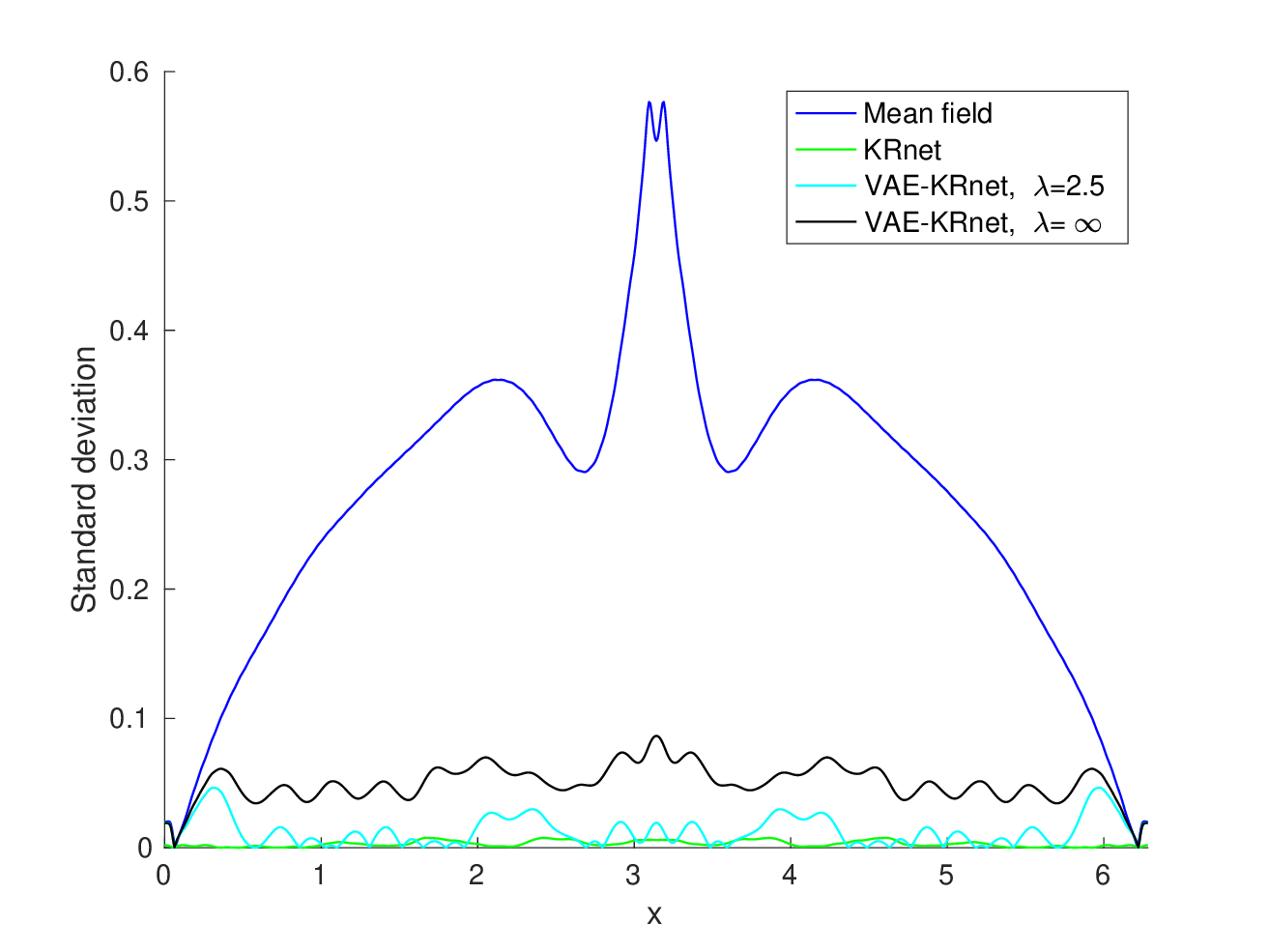}}
	\caption{Errors of different models on the validation set. The errors are scaled by $\|\mathbb{E}[r]\|_{L2}$. $n=50$. $d=8$. Left: mean; Right: variance.}\label{fig:d50_II_mean_var}
\end{figure}

We subsequently look at the case that $n=50$, where we change $\alpha$ in $b_{ij}=e^{-|i-j|/\alpha}$ from 3 to 10. So more correlations can be introduced such that a relatively small number of latent random variables is needed. For this case, $-\log C=632.11$. We let $d=8$. The global evolution behavior of the ADAM method is similar to the case that $n=10$, see figure \ref{fig:d50_II_cong}. Note that although the number of dimensions is relatively large, the KRnet accurately captured the correlations using a model that has the same complexity (in terms of $K$, $L$ and $N_L$) as the model for the case $n=10$. We plot pointwise errors in terms of $x$ in figure \ref{fig:d50_II_mean_var} for the mean on the left and for the standard deviation on the right. Both errors are scaled by the $L_2$ norm of the exact mean, i.e., $\|\mathbb{E}[r]\|_{L_2}$. For this case, the mean-field model yields the best estimation for the mean but no useful estimation for the standard deviation. KRnet yields accurate predictions for both the mean and the standard deviation. VAE-KRnet with $\lambda=\infty$ yields a better estimation for the mean and a worse estimation for the variance than VAE-KRnet with $\lambda=2.5$. The rank for predicting the mean is: mean-field model, KRnet, VAE-KRnet with $\lambda=\infty$, VAE-KRnet with $\lambda=2.5$. The rank for predicting the variance is: KRnet, VAE-KRnet with $\lambda=2.5$, VAE-KRnet with $\lambda=\infty$ and mean-field model. 

\begin{figure}	
	\center{
		\includegraphics[width=0.6\textwidth]{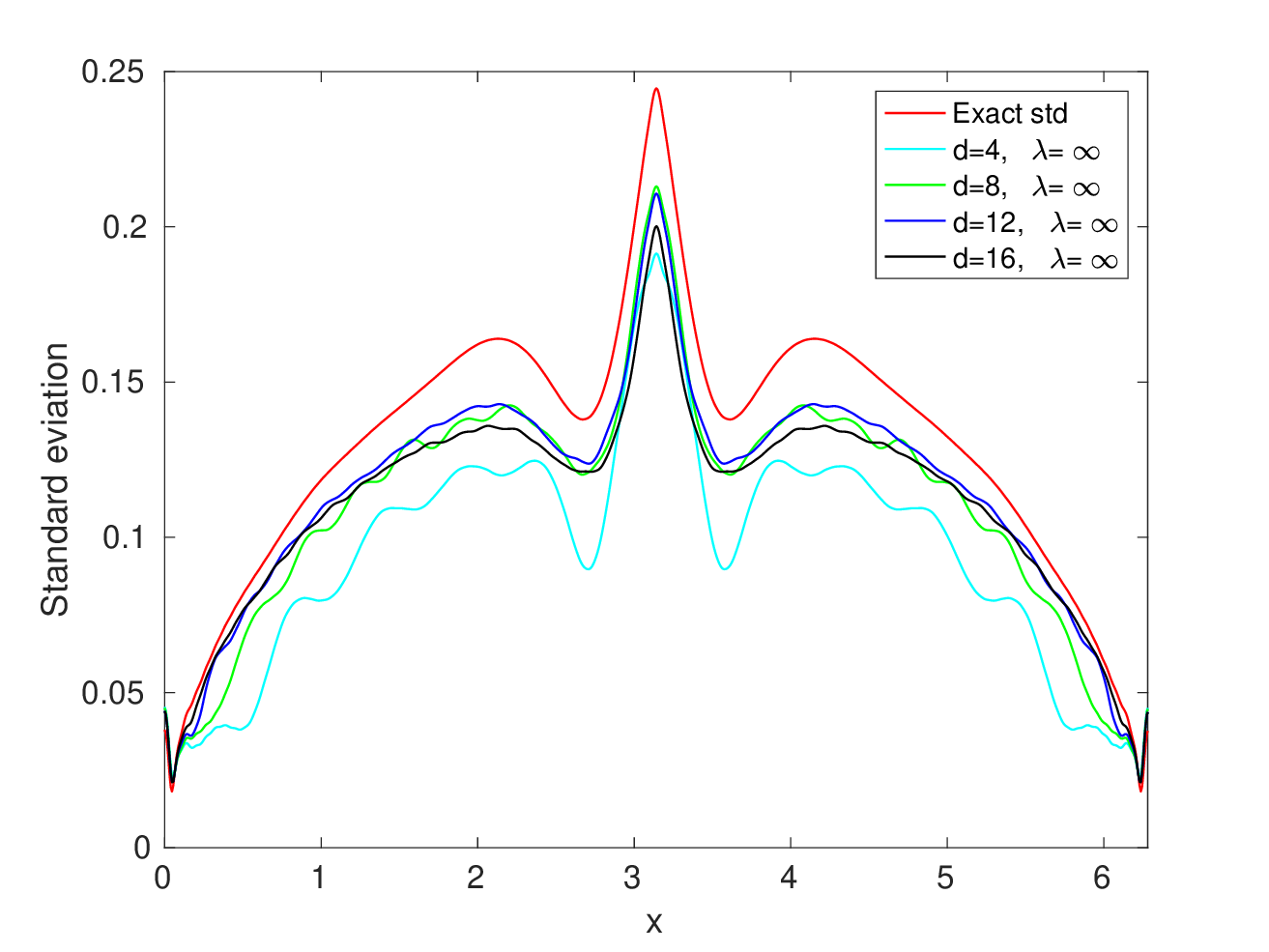}}
	\caption{Standard deviation given by VAE-KRnet with varying $d$. $n=50$.}\label{fig:d50_II_mean_var_mx}
\end{figure}

We finally compare the predictions given by VAE-KRnet in terms of the dimension of the latent variables. The results are plotted in figure \ref{fig:d50_II_mean_var_mx}. For a certain $d$, we choose $K=d/4$ for KRnet, i.e., the dimensions will be deactivated by $d/4$. This way, the overall number of model parameters of VAE-KRnet remains almost the same for a varying $d$. It is seen that as $d$ increases the improvement on the prediction will cease at a certain $d$. The reason is twofold: first, VAE does not converge to the full model when $d$ increases to $n$ because of the model error; second, we constrained the model complexity with a roughly constant number of model parameters.  

\subsection{One-dimensional elliptic problem}
\begin{figure}	
\center{
\includegraphics[width=0.49\textwidth]{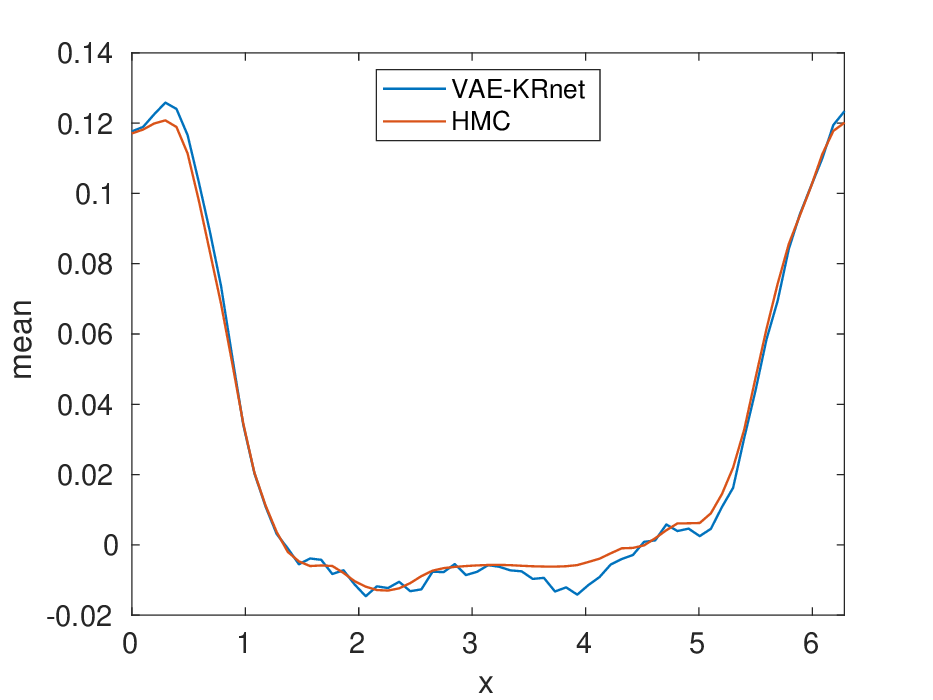}
\includegraphics[width=0.49\textwidth]{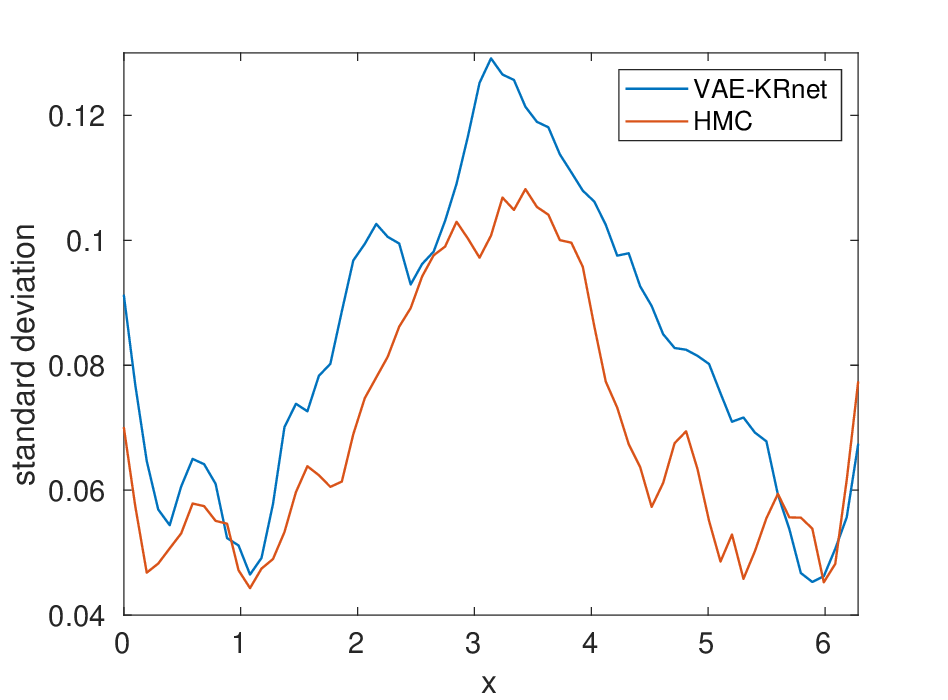}}
	\caption{Statistics of $\mathbf{a}$ given by VAE-KRnet and HMC for the elliptic problem. $n=129$. $d=16$. Left: mean; Right: standard deviation.}\label{fig:mean_var_vae_hmc}
\end{figure}

{
We now consider a one-dimensional elliptic problem 
\begin{equation}
-\frac{d}{dx}\left(e^{a(x;\omega)}\frac{du}{dx}\right)=1,\quad x\in[0,1],
\end{equation}
with homogeneous boundary conditions, where $a(x;\omega)$ is a Gaussian field. We associate $a(x;\omega)$ with a Gaussian measure $\mu=\mathcal{N}(0,\mathcal{C})$ with zero mean and covariance operator $\mathcal{C}$ in $L^2([0,1])$. The covariance operator $\mathcal{C}$ is chosen as
\begin{equation}
\mathcal{C}=\alpha^{-1}(I-\Delta)^{-s}=\alpha^{-1}\mathcal{A}^{-s},
\end{equation}
where $\alpha>0$ is a constant, $s>\frac{1}{2}$, and the domain of $\mathcal{A}$ is 
\begin{equation}
D_{\mathcal{A}}=\{a(x)\in H^2([0,1]):\partial_xa=0\textrm{ at }x=0,1\}.
\end{equation}
We infer the coefficient $a(x)$ using pointwise observations of $u(x)$:
\begin{equation}
\hat{u}_i=u(x_i)+\eta_i,\quad i=1,\ldots,J,
\end{equation}
where $J$ is the total number of observation locations $x_i$ and $\eta_i\sim\mathcal{N}(0,\sigma^2)$ are i.i.d. Gaussian random variables with zero mean and variance $\sigma^2$. In the framework of Bayesian inverse problem, the posterior measure $\nu$ can be represented with the Radon-Nikodym derivative
\begin{equation}
\frac{d\nu}{d\mu}(a(x)|\hat{\bu})\propto \exp(-\Phi(a,\hat{\bu})),
\end{equation}
where 
\begin{equation}
\Phi(a,\hat{\bu})=\frac{1}{2\sigma^2}\sum_{i=1}^K(\hat{u}_i-u(x_i))^2
\end{equation}
is the potential defined by the distribution of the observational noise $\eta_i$. 

We will approximate the posterior distribution using VAE-KRnet. We discretize the problem using linear finite elements given by $\mathrm{span}\{\theta_i(x)\}_{i=1}^M$. Let $\mbA$ be the representation of $\mathcal{A}$ in the finite element space, i.e.,
\begin{equation}
\mbA=\mbM^{-1}\mbK+\mbI,
\end{equation}
where $\mbM$ is the mass matrix and $\mbK$ is the stiffness matrix. Let $(\sigma_i,\bv_i)$ be the eigenpairs of $\mbA$, i.e.,
\begin{equation}
\mbA\bv_i=\sigma_i\bv_i,\quad \mbA\mathbf{\mathbf{V}}=\mathbf{V}\mathbf{\Sigma},\quad\bv_i^\mathsf{T}\mbM\bv_j=\delta_{ij},
\end{equation}
where $\mathbf{\Sigma}$ is a diagonal matrix with nonzero entries $\sigma_i$ and $\mathbf{\Lambda}$ include all eigenvectors $\bv_i$. Using the matrix transfer technique, we may define the discrete representation of $\mathcal{A}^{-s}$ as
\begin{equation}
\mbA^{-s}=\mathbf{V}\mathbf{\Sigma}^{-s}\mathbf{V}^{-1},
\end{equation}
where 
\begin{equation}
\mathbf{A}^{-s}\bv_i=\sigma_i^{-s}\bv_i=\lambda_i\bv_i.
\end{equation}
We then have the finite element representation of $a(x)=\ba^{\mathsf{T}}\mathbf{\Theta}(x)$ using the following Karhunen-Loeve expansion
\begin{equation}
\ba=\ba_0+\mathbf{V}\mathbf{\Lambda}^{1/2}\bf{\bgamma},
\end{equation}
where $\bgamma\in\mathcal{N}(0,\mathbf{I})$, $\ba\in\mathbb{R}^M$ includes the coefficients of the finite element approximation, $\mathbf{\Theta}(x)\in\mathbb{R}^M$ includes all finite element basis functions and $\mathbf{a}_0$ is the mean. With respect to $\ba$, we have the discretized posterior distribution
\begin{equation}
\pi(\ba|\hat{\bu})\propto\exp\left(-\Phi(\ba,\hat{\bu})\right)\mathcal{N}(\ba_0,\mathbf{V}\mathbf{\Lambda}\mathbf{V}^{\mathsf{T}}),
\end{equation}
where $\Phi(\ba,\hat{\bu})$ is the discretized potential induced by $\ba$. Note that the dimension of $\ba$ is the number of finite element basis functions. 

For this case explicit forms are not available for the posterior measure $\nu$ or its discrete version $\pi(\ba|\hat{\bu})$. We here use the function-space hybrid Monte Carlo (HMC) to generate some reference solutions \cite{Beskos_2011}. We use $M=128$ equidistant linear finite elements to discretize $[0,1]$. The observations are generated with $a(x)=\exp(0.1\cos(2\pi x))$ on locations $x_i=\frac{i}{64}$, $i=1,\ldots,63$, where the noise amplitude $\sigma$ is $5\%$ of  the maximum of $u(x)$. We let $\lambda=2.8$ in equation \eqref{eqn:PDF_vae_krnet_mutual} according to the right plot of figure \ref{fig:d10_II_mean_var_mx}. The dimension of the latent space is set to be $d=16$ in contrast to the dimension $n=M+1=129$ of $\ba$. 

We use the same configuration as the case in the previous section for the definition of VAE-KRnet. The only difference is that in KRnet the number of active dimensions will be deactivated by $16/4=4$. In each optimization step, we need to compute $\nabla_{\ba}\pi$, where one forward problem and one adjoint problem are solved, and other gradients can be done by the automatic differentiation in Tensorflow. Due to the cost of solving two elliptic problems in each iteration, we consider a relatively small training set of size 1024 with a batch size 128. In contrast to the hybrid Monte Carlo, the computation of the forward and adjoint problems can be parallelized when training VAE-KRnet. We train the model for 600 epochs and use the models given by the last 60 epochs for a time average when computing predictions. For the hybrid Monte Carlo, we generated $10^5$ samples. For this case, the time consumption for VAE-KRnet is about one fourth of that for HMC.

In figure \ref{fig:mean_var_vae_hmc} we plotted some statistics of $\ba$ in terms of the posterior distribution. It is seen that VAE-KRnet provides consistent results with HMC for both the mean and the standard deviation although the size of the training set is relatively small. We here simply use the one-dimensional elliptic problem to verify the effectiveness of VAE-KRnet. Many computational issues remain to improve the efficiency of VAE-KRnet, which are beyond the scope of this paper.
}

\section{Summary}
In this paper, we have developed a family of probability density models by coupling VAE and KRnet. KRnet is an effective invertible transport map, and VAE is an effective technique for dimension reduction.  VAE-KRnet inherits the advantages of both VAE and KRnet. For a linear system, the encoder of the canonical VAE mainly does the following things (see Remark \ref{rmk:encoder}): 1) A mapping from a standard Gaussian to an arbitrary distribution in the latent space; 2) A rotation to make the encoder diagonal; and 3) computation of the covariance matrix; In the canonical VAE, all these tasks are achieved by the encoder. In VAE-KRnet, the first task is achieved by a KRnet for an arbitrary prior, and the second task is achieved by another KRnet for the correlation between dimensions.  Compared to the canonical VAE, each component of VAE-KRnet has a more specific task, which improves both the performance and the robustness. We applied VAE-KRnet to variational Bayes to approximate the posterior. VAE-KRnet has demonstrated some promising potentials: 1) It covers a wide range of data dimensions by varying the number of dimensions of the latent space from zero (KRnet) to $d$ (VAE-KRnet). 2) By taking into account the mutual information, a possibility is provided to improve the underestimation of the variance by varying the parameter $\lambda$, which yields a statistics-oriented way for model selection. 3) Increasing the dimensionality $d$ will improve the approximation. Of course, $d$ would be limited by the model capability of both VAE and KRnet as shown in figure \ref{fig:d50_II_mean_var_mx}. However, varying $d$ does not need to introduce a significant change of the model complexity. 4) For linear Bayesian inverse problems, the KRnet performs very well for high-dimensional cases. It is seen that KRnet yields the best prediction for the last example with $n=50$. On one hand,  this is because the posterior is Gaussian, which is relatively simple; on the other hand, it demonstrates the modeling capability of KRnet. 5) The numerical experiments on the elliptic problem demonstrate that VAE-KRnet is able to produce consistent results with HMC using a relatively small training set. The current results are very encouraging for us to apply VAE-KRnet to model the posterior of nonlinear Bayesian inverse problems. 

\section*{Acknowledgment}
X. Wan has been supported by NSF grant DMS-1913163, and S. Wei has been supported by NSF grant ECCS-1642991.

\end{document}